\documentclass[11pt,twoside]{article}
\usepackage[a4paper,top=1in,bottom=1in,left=0.8in,right=0.8in]{geometry}

\usepackage[lining,proportional]{ebgaramond}
\newcommand{\headingfont}{\normalfont\scshape}

\usepackage{fancyhdr}
\usepackage{titlesec}
\usepackage{enumitem}
\usepackage[round,authoryear]{natbib}

\IfFileExists{math_commands.tex}{%%%%% NEW MATH DEFINITIONS %%%%%

\usepackage{amsmath,amsfonts,bm}

\def\eqref#1{equation~\ref{#1}}
\def\1{\bm{1}}

\DeclareMathAlphabet{\mathsfit}{\encodingdefault}{\sfdefault}{m}{sl}
\SetMathAlphabet{\mathsfit}{bold}{\encodingdefault}{\sfdefault}{bx}{n}

\newcommand{\E}{\mathbb{E}}

\newcommand{\R}{\mathbb{R}}

}{%
  \newcommand{\E}{\mathbb{E}}%
  \newcommand{\R}{\mathbb{R}}%
}

\usepackage[utf8]{inputenc}
\usepackage[T1]{fontenc}
\usepackage{hyperref}
\usepackage{url}
\usepackage{booktabs}
\usepackage{amsmath,amssymb,amsfonts,amsthm}
\usepackage{mathtools}
\usepackage{bm}
\usepackage{nicefrac}
\usepackage{microtype}
\usepackage[dvipsnames]{xcolor}
\usepackage{graphicx}
\usepackage{subcaption}
\usepackage{wrapfig}
\usepackage{physics}

\newcommand{\preprintgraphic}[2][]{%
  \IfFileExists{#2}{\includegraphics[#1]{#2}}{%
    \fbox{\parbox[c][34mm][c]{0.9\linewidth}{\centering\small
      Figure asset: \texttt{\detokenize{#2}}}}}}

\setlist[itemize]{leftmargin=1em,itemsep=0.1em,topsep=0.45em}

\newcommand{\vA}{A}
\newcommand{\cN}{\mathcal{N}}

\newcommand{\bgamma}{\bm{\gamma}}
\newcommand{\bh}{\bm{h}}
\newcommand{\bw}{\bm{w}}
\newcommand{\bx}{\bm{x}}
\newcommand{\ba}{\bm{a}}
\newcommand{\bv}{\bm{v}}
\newcommand{\x}{\bm{x}}
\newcommand{\h}{\bm{h}}
\newcommand{\task}{\bm{w}}
\newcommand{\prox}{\operatorname{prox}}
\newcommand{\iid}{\stackrel{\mathrm{i.i.d.}}{\sim}}
\DeclareMathOperator{\vecop}{vec}

\usepackage[most]{tcolorbox}
\definecolor{CiteColor}{HTML}{0A6E3A}
\definecolor{SideBarGreen}{HTML}{1A4B31}
\definecolor{TextDarkGreen}{HTML}{1A4B31}
\definecolor{GreenBorder}{HTML}{A0B8A9}
\definecolor{GreenPanel}{HTML}{F5FAF7}
\definecolor{SideBarBrown}{HTML}{613D04}
\definecolor{TextDarkBrown}{HTML}{362205}
\definecolor{BrownBorder}{HTML}{C2B8A5}
\definecolor{BrownPanel}{HTML}{F7F5F2}
\definecolor{PreprintGreen}{HTML}{184B38}
\definecolor{PreprintMuted}{HTML}{5B665F}
\tcbset{
  resultbox/.style={
    enhanced,
    breakable,
    sharp corners,
    boxrule=0.55pt,
    opacityback=1,
    boxsep=0pt,
    left=11pt,
    right=11pt,
    top=9pt,
    bottom=9pt,
    before skip=10pt,
    after skip=10pt,
    fonttitle=\bfseries,
    detach title,
  },
  greenpalette/.style={
    colframe=GreenBorder,
    colback=GreenPanel,
    borderline west={1.6pt}{0pt}{SideBarGreen},
    coltitle=TextDarkGreen,
    before upper={
      \noindent{\color{SideBarGreen}\bfseries\tcbtitle}\par
      \vspace{0.35em}
      {\color{GreenBorder!45}\hrule height 0.35pt}
      \vspace{0.75em}
    },
  },
  brownpalette/.style={
    colframe=BrownBorder,
    colback=BrownPanel,
    borderline west={1.6pt}{0pt}{SideBarBrown},
    coltitle=TextDarkBrown,
    before upper={
      \noindent{\color{SideBarBrown}\bfseries\tcbtitle}\par
      \vspace{0.35em}
      {\color{BrownBorder!45}\hrule height 0.35pt}
      \vspace{0.75em}
    },
  },
}

\newenvironment{boxedproposition}[1]{%
  \refstepcounter{proposition}%
  \begin{tcolorbox}[
    resultbox,
    greenpalette,
    title={Result~\theproposition: #1}
  ]%
}{%
  \end{tcolorbox}%
}
\newenvironment{boxedlemmabrown}[1]{%
  \refstepcounter{proposition}%
  \begin{tcolorbox}[
    resultbox,
    brownpalette,
    title={Result~\theproposition: #1}
  ]%
}{%
  \end{tcolorbox}%
}

\hypersetup{colorlinks=true,linkcolor=CiteColor,citecolor=CiteColor,urlcolor=CiteColor}

\titleformat{\section}
  {\Large\headingfont\color{black}}
  {\thesection}{0.65em}{}
\titleformat{\subsection}
  {\large\headingfont\color{black}}
  {\thesubsection}{0.65em}{}
\titleformat{\subsubsection}
  {\normalsize\headingfont\color{black}}
  {\thesubsubsection}{0.65em}{}
\titleformat{\paragraph}[runin]
  {\normalsize \bfseries \color{black}}
  {}{0pt}{}[.]
\titlespacing*{\section}{0pt}{2.2ex plus .5ex minus .2ex}{1.0ex}
\titlespacing*{\subsection}{0pt}{1.8ex plus .4ex minus .2ex}{0.7ex}

\newcommand{\shortauthors}{Rysmakhanov, Letey, Lu, Zavatone-Veth \& Pehlevan}
\makeatletter
\renewenvironment{abstract}{%
  \begin{center}
  \begin{minipage}{0.91\textwidth}
  \noindent{\headingfont\color{black}Abstract}\par\smallskip
}{%
  \end{minipage}
  \end{center}
  \vspace{0.8em}
}
\makeatother

\def\And{%
  \end{tabular}\hfil\linebreak[0]\hfil
  \begin{tabular}[t]{c}\ignorespaces
}
\def\AND{%
  \end{tabular}\hfil\linebreak[4]\hfil
  \begin{tabular}[t]{c}\ignorespaces
}

\newcommand{\hataone}{\widehat a_1}
\newcommand{\hatatwo}{\widehat a_2}
\newcommand{\hatazero}{\widehat a_0}
\newcommand{\aone}{a_1}
\newcommand{\atwo}{a_2}
\newcommand{\azero}{a_0}
\newcommand{\fitlogit}[1]{R_{#1}}

\title{\vspace{-1.25em}{\LARGE\headingfont\color{black}
In-context classification: Calibration and noise \\ in linear attention with softmax readout}}

\author{%
  Arman Rysmakhanov\textsuperscript{$\ast$,\textdagger,1,2} \\
  {\small\href{mailto:ar21@williams.edu}{\texttt{ar21@williams.edu}}}
  \And
  Mary Letey\textsuperscript{$\ast$,1,3} \\
  {\small\href{mailto:maryletey@fas.harvard.edu}
  {\texttt{maryletey@fas.harvard.edu}}}
  \AND
  Yue M.~Lu\textsuperscript{1} \\
  {\small\href{mailto:yuelu@seas.harvard.edu}
  {\texttt{yuelu@seas.harvard.edu}}}
  \And
  Jacob A.~Zavatone-Veth\textsuperscript{4} \\
  {\small\href{mailto:jzavatoneveth@fas.harvard.edu}
  {\texttt{jzavatoneveth@fas.harvard.edu}}}
  \And
  Cengiz Pehlevan\textsuperscript{1,3} \\
  {\small\href{mailto:cpehlevan@seas.harvard.edu}
  {\texttt{cpehlevan@seas.harvard.edu}}}
}
\date{}

\begin{document}

\maketitle
\vspace{-2em}

{\centering\small
  \textsuperscript{1}Applied Mathematics, Harvard University
  \hspace{1.5em}
  \textsuperscript{2}Williams College
  \hspace{1.5em}
  \textsuperscript{3}Kempner Institute, Harvard University
  \\[0.15em]
  \textsuperscript{4}Center for Brain Science and Society of Fellows,
  Harvard University
  \par
}

\begingroup
  \renewcommand{\thefootnote}{\fnsymbol{footnote}}
  \footnotetext[1]{Equal contribution.}
  \footnotetext[2]{This work was performed while at Harvard University.}
\endgroup

\thispagestyle{empty}

\vspace{0.8em}

\begin{abstract}
We study in-context binary classification in a simple attention model trained to minimise a logistic loss. Each prompt has a latent logistic teacher, and the learner must infer its direction from Bernoulli-labeled context examples. We derive an asymptotic prediction for the test logit in a high-dimensional regime with context length $\ell \propto d$ and total training prompts $n \propto d^2$. Our formula cleanly separates calibration, test-time context noise, and non-isotropic parameter-estimation noise as interpretable effects on in-context performance. Our analysis also identifies the ridgeless separability regime and addresses the effects of context length, pretraining-set size, regularisation, and sampling temperature on in-context classification performance. The theory's predictions closely agree with finite-dimensional simulations and fully trained linear-attention models, while experiments with standard softmax attention show qualitatively related behaviour beyond the regime assumed by our analysis.
\end{abstract}

\section{Introduction}
In-context learning (ICL) is the ability of a model to adapt its predictions from examples in a prompt, without updating its parameters \citep{brown2020language, vonoswald2023transformers, wei2022emergent}. It is a striking capability of modern sequence models, which can, for instance, implement a range of statistical algorithms in-context \citep{bai2023transformers}. However, the statistical mechanism that turns a context into a new-task prediction is not fully understood.

Simplified ICL models offer a useful route to this question because they make explicit which prompt statistics an architecture can use \citep{akyurek2023what,vonoswald2023transformers,zhang2023doesincontextlearninglearn,lu2025asymptotictheoryincontextlearning, oko2024pretrainedtransformerefficientlylearns,vasudeva2025transformers,letey2025pretrain,letey2026sequential}. In particular, a significant body of recent work has explored ICL in the context of the test problem of linear regression. For these tasks, linear attention \citep{katharopoulos2020transformers} has been shown to be a good surrogate for understanding how and what transformers learn \citep{ahn2024linear,zhang2023trained, wu2023pretraining, lu2025asymptotictheoryincontextlearning, letey2025pretrain,letey2026sequential, bordelon2025theoryscalinglawsincontext}. Together, these works demonstrate how reduced models can yield substantial insight into the mechanisms of ICL.

Despite extensive progress in the theory of in-context learning, many works focus on regression; binary classification introduces qualitatively different questions and phenomena. In the setting studied here, each prompt is generated by a latent task direction, and its context labels are Bernoulli observations determined by that direction. A successful predictor must therefore output a calibrated probability rather than merely recover a sign. Logistic regression, a canonical high-dimensional inference problem, is a natural target for such investigation. This problem has been widely studied \citep{mai_largescale_logistic_asymp,Sur_2019}, including its phase transitions \citep{candes2018phasetransitionexistencemaximum,barbier2019optimal}, universality \citep{pesce2023gaussiandataneedextents}, and generalisation \citep{gerace2020generalisation,mei2022generalization}, leading to a rich understanding of a single instance of this estimation problem.

By contrast, the in-context setting involves learning a general estimator that can approximately solve new instances of logistic regression. Recent works have investigated whether transformers can learn to do this, perhaps by performing normalised gradient descent \citep{zhang2026transformers}. Moreover, studies of multiclass classification have shown that properties of the data greatly affect the model's ability to learn in-context \citep{chan2022data}. Yet, a detailed understanding of in-context classification, comparable to what has been achieved for regression, is still lacking.

Here, we study in-context logistic regression using a reduced linear-attention model with a softmax readout. We derive a high-dimensional characterisation of the output logit, and corresponding test errors, in a proportional regime where context length scales with token dimension and the number of pretraining sequences scales with its square. We characterise the joint law of the output logit and teacher logit, leading to the following findings. \begin{itemize}
    \item The form of the output logit is highly interpretable, and decomposes into three parts: scaling of the teacher-aligned signal, noise from test-time context, and anisotropy induced by finitely many pretraining sequences.
    \item This separation distinguishes the roles of context length and training-set size, and recovers the Bayes logit in the joint large-context, large-pretraining-data ridgeless limit.
    \item We predict the separability boundary at which the ridgeless logistic estimator ceases to have a finite solution.
    \item Finally, we verify the theory through finite-dimensional simulations and show that its predicted logit distribution describes the output of a trained linear-attention model.
\end{itemize}

\section{Data and model setup}

\paragraph{Data generation} Following prior theoretical work on ICL \citep{zhang2023trained,wu2023pretraining,lu2025asymptotictheoryincontextlearning}, we train the model on a dataset of $n$ sequences of examples. Each sequence consists of $(x,y)$ pairs sampled from a latent task, and the model aims to learn the corresponding conditional distribution. Each sequence has a latent task vector $\bw\in\R^d$ drawn uniformly from the unit sphere. Fix an inverse temperature $\beta>0$. Fixing $\bw$, draw independent covariates $\bx_i\sim\cN(0,I_d)$ and a query $\bx_\mathrm{q}\sim\cN(0,I_d)$, with
\begin{equation}
 y_i\mid \bx_i,\bw\sim\operatorname{Bernoulli}\!\left(\sigma(\beta\bw^\top\bx_i)\right),
 \qquad
 y_q\mid \bx_\mathrm{q},\bw\sim\operatorname{Bernoulli}\!\left(\sigma(\beta\bw^\top\bx_\mathrm{q})\right),
 \label{eq:teacher}
\end{equation}
where $\sigma(t)=(1+e^{-t})^{-1}$ is the sigmoid function.

\paragraph{Model and estimator}
Following a standard sequence-embedding convention used in prior works \citep{vonoswald2023transformers, zhang2023trained, lu2025asymptotictheoryincontextlearning}, we embed each sequence as a matrix
\begin{equation*}
 Z=\begin{bmatrix}\bx_1&\cdots&\bx_\ell&\bx_\mathrm{q}\\ y_1&\cdots&y_\ell&0\end{bmatrix}\in\R^{(d+1)\times(\ell+1)}.
\end{equation*}
The final zero is a placeholder for the query label $y_\mathrm{q}$. We pass $Z$ to a linear-attention operator and map the final, label-coordinate entry of the updated query column to a binary class probability. Specifically, writing $\bm{z}_i=[\bx_i^\top\ y_i]^\top$ and $\bm{z}_\mathrm{q}=[\bx_\mathrm{q}^\top\ 0]^\top$, the query column updated by linear attention is
\vspace{-0.2cm}
\begin{equation}
  \operatorname{LA}_{Q,K,V}(Z)_{:,\ell+1}
 =\bm{z}_\mathrm{q}+\frac1\ell\sum_{i=1}^\ell
 \langle Q\bm{z}_\mathrm{q},K\bm{z}_i\rangle V\bm{z}_i.
 \label{eq:linear-attention}
\end{equation}
\vspace{-0.3cm}

Here $Q,K,V\in\mathbb{R}^{(d+1)\times(d+1)}$ are the query, key, and value matrices, respectively, that parameterise linear attention. Since this is binary classification, the softmax depends only on the difference between the output logits. We use the final, label entry as the class-1 logit and fix the class-0 logit to zero,
\begin{align}
r_{Q,K,V}(Z)&=\big[\operatorname{LA}_{Q,K,V}(Z)\big]_{d+1,\ell+1}\\
 \text{Out}(Z)&=\big[\operatorname{softmax}(0,r_{Q,K,V}(Z))\big]_2
 =\sigma(r_{Q,K,V}(Z)).
 \label{eq:linear-attention-logit}
\end{align}

Prior work on in-context linear regression has shown that linear attention admits a particularly clean reduced representation: by expanding (\ref{eq:linear-attention-logit}) in terms of $V$ and $Q^\top K$, not all components of these matrices contribute to learnable directions, and can be ignored without meaningfully changing the optimisation dynamics \citep{zhang2023trained,lu2025asymptotictheoryincontextlearning}. We utilise this established reduction to simplify the output logit,
\begin{align}
r_{Q,K,V}(Z) &\approx \bx_\mathrm{q}^\top\left(A\frac d\ell\sum_{i=1}^\ell y_i\bx_i
 +\ba\frac1\ell\sum_{i=1}^\ell y_i\right) \\
A&=\frac{1}{d}V_{d+1,d+1}(Q^\top K)_{1:d,1:d}\,, \quad \bm{a}^\top = V_{d+1,d+1}(Q^\top K)_{1:d,d+1}^\top\,.
\end{align} We will later verify this by directly comparing our results for this reduced model to a fully parameterised, trained linear-attention model. We can write the reduced model more compactly by introducing the new parameter matrix $\Gamma = [A \;\; \bm{a}] \in \mathbb{R}^{d\times (d+1)}$ and
\begin{equation}
r_\Gamma(Z)=\langle\Gamma,H\rangle, \quad H=\bx_\mathrm{q}\begin{bmatrix}\bv^\top&c\end{bmatrix}, \quad \bv=\frac d\ell\sum_{i=1}^{\ell}y_i\bx_i, \quad c=\frac1\ell\sum_{i=1}^{\ell}y_i.
 \label{eq:logit}
\end{equation}
Finally, applying softmax gives the class-probability estimator $\text{Out}_\Gamma(Z)=\sigma(r_\Gamma(Z))$.

\paragraph{Optimal parameters given training batch} We consider a training batch with $n$ independent prompts, which yields $n$ independent feature matrices $H_\mu$ ($\mu\in[n]$) in the reduced logit $r_\Gamma$. Write $\bgamma=\vecop(\Gamma)$ and $\bh_\mu=\vecop(H_\mu)$ for their row-wise vectorisations in $\mathbb{R}^{d(d+1)}$. We use $y_\mu$ as shorthand for the $\mu^\mathrm{th}$ sequence's query label $y_\mathrm{q}$, and fit the reduced model by regularised Bernoulli negative log-likelihood:
\vspace{-0.3cm}
\begin{equation}
 \widehat\bgamma=\arg\min_{\bgamma}\left\{
 \frac1n\sum_{\mu=1}^n\big[L(\bgamma^\top\bh_\mu)-y_\mu\bgamma^\top\bh_\mu\big]
 +\frac{\lambda d}{2}\|\bgamma\|^2\right\},
 \quad L(t)=\log(1+e^t).
 \label{eq:objective}
\end{equation}
The choice of $\lambda d$ scaling ensures that the regularisation is applied evenly and consistently across dimensions, as measured by comparison to the empirical Hessian trace: the average Hessian eigenvalue grows with $d$ and so without this normalisation, the effective regularisation would be zero at large $d$.

\paragraph{Asymptotic assumptions} We study the proportional limit $n,\ell,d\to\infty$ with constant
\[
\tau := \frac{n}{d^2}, \qquad \alpha_\mathrm{train}:=\frac{\ell}{d}.
\]

\section{High-dimensional characterisation}
For the fitted, data-dependent parameters $\widehat\Gamma=[A^*\;\bm a^*]$, define the empirical order parameters
\begin{equation}
 \hataone=\tr(A^*),\qquad \hatatwo=d\|A^*\|_F^2,\qquad \hatazero=\|\ba^*\|^2.
 \label{eq:orders}
\end{equation}
We will derive limiting deterministic values for these random parameters. The parameter $\hatazero$ measures the strength of the vector component of $\widehat\bgamma$. For the other two, note that
\vspace{-0.2cm}
$$ \hatatwo-\hataone^2=d\|A^*\|_F^2-\bigl(\tr A^*\bigr)^2
 =d\left\|A^*-\frac{\tr(A^*)}{d}I_d\right\|_F^2\geq 0. $$
This quantity is minimised at $A^*\propto I_d$, so it measures the squared distance of $A^*$ from isotropy.

We aim to derive a statistically equivalent expression for the test logit, where $\bm{h}_\mathrm{test}$ is drawn fresh from the distribution described above, and the optimal parameters $\widehat{\bm{\gamma}}$ are set from a finite training set of sequences as above. We highlight that this will provide richer phenomenology than an analysis of the population optimal parameters, which we begin with to provide intuition.

\subsection{Population limits}
Consider $\Gamma_\mathrm{pop} = [{A}^*_\mathrm{pop}\; \bm{a}^*_\mathrm{pop}]$ and $\bm{\gamma}_\mathrm{pop} = \vecop(\Gamma_\mathrm{pop})$ defined by \begin{equation}
 \bm{\gamma}_\mathrm{pop} =\arg\min_{\bgamma}\left\{
 \E_{\bm{h},y_\mathrm{q}} \big[L(\bgamma^\top\bh)-y_\mathrm{q}\bgamma^\top\bh\big]
 +\frac{\lambda d}{2}\|\bgamma\|^2\right\} = \arg\min_{\bgamma} \mathcal{L}_\mathrm{pop}(\bgamma).
 \label{eq:population_objective}
\end{equation}
For $\lambda>0$, this objective is strictly convex in $\bm{\gamma}$ and therefore has a unique minimiser. The linear features $\bw^\top\bm{x}$ are rotationally invariant, since $(O\bw)^\top(O\bm{x})=\bw^\top\bm{x}$ for every orthogonal $O$. The structure of the linear-attention estimator then gives
\[\mathcal{L}_\mathrm{pop}(\text{vec}([A\;\bm{a}]))=\mathcal{L}_\mathrm{pop}(\text{vec}([OAO^\top\;O\bm{a}]))
\]
for \textit{every} orthogonal $O$. Uniqueness of the population solution therefore implies $A^*_{\mathrm{pop}}\propto I_d$ and $\bm{a}^*_{\mathrm{pop}}=\bm{0}$. Thus isotropy emerges as $\tau\to\infty$, whereas the finite-sample result can retain an anisotropic contribution.

We can separately consider the long-context limit. Holding $n$ and $d$ fixed, take $\ell\to\infty$. Then
\vspace{-0.2cm}
\begin{equation}
\frac1d\bv=\frac1\ell\sum_{i=1}^\ell y_i\bx_i
 \longrightarrow \mathcal{C}_\beta\bw,
 \qquad
 c = \frac{1}{\ell}\sum_{i=1}^\ell y_i \longrightarrow \frac12,
 \label{eq:infinite-context-statistics}
\end{equation}
where $\mathcal{C}_\beta:=\E_{G\sim\cN(0,1)}[G\sigma(\beta G)]
 =\beta\E[\sigma'(\beta G)]$ is the first Hermite coefficient of the sigmoid activation. Thus the reduced-model logit becomes
\begin{equation} r_\Gamma(Z)\longrightarrow
 d\mathcal{C}_\beta\bx_\mathrm{q}^\top A\bw
 +\frac12\bx_\mathrm{q}^\top\ba.
 \label{eq:infinite-context-logit}
\end{equation}

We see that these two limits remove different sources of randomness. The large-$n$ limit removes randomness from finite-sequence anisotropy, while the long-context limit removes randomness due to misestimation of the task in each sequence. Only in the jointly long-context, population-training limit do we see both of these effects disappearing; in the ridgeless case, this amounts to a recovery of the Bayes logit $r(Z) \to \beta G_\mathrm{q}$ for $G_\mathrm{q}\sim \mathcal{N}(0,1)$. We will see such different sources of randomness in the finite-sample calculation below.

\subsection{Finite-sample results: order parameters} We derive an equivalent distribution for the test logit, using fitted parameters $\widehat{\bgamma}=\vecop(\widehat{\Gamma})$. The full derivation and its assumptions are given in Appendix~\ref{app:derivation}. Schematically, we use a cavity approach \citep{el2018impact}: we exclude one of the $n$ prompts from the loss and study the effect of this omission on the fitted parameters. In particular, we study the deterministic limits $\aone,\atwo,$ and $\azero$ of the empirical order parameters $(\hataone,\hatatwo,\hatazero)$, which characterise properties of the optimised weights $\widehat{\bgamma}$.

\begin{boxedlemmabrown}{Self-consistency equations for order parameters}\label{lemma:a1a2eqs}
The order parameters are determined implicitly as follows. Fix $\alpha_\mathrm{train}, \tau, \beta$ and $\lambda$. For any candidate pair $(\aone,\atwo)$ such that $\atwo \geq \aone^2$, draw independent $G_\mathrm{q},G_{\rm tr},G_\perp\sim\cN(0,1)$ and then draw from the label distribution $Y\mid G_\mathrm{q}\sim\operatorname{Bernoulli}(\sigma(\beta G_\mathrm{q})).$ Define $\mathcal{C}_\beta = \E_{g\sim \mathcal{N}(0,1)}[g\sigma(\beta g)]\,, \Delta = \mathcal{C}_\beta^2 +1/({2\alpha_\mathrm{train}}).$
Use the candidate order parameters to form the trained
logit $R_s$ as the unique solution of
\begin{equation}
R_s+s\big(\sigma(R_s)-Y\big) =  \aone\mathcal{C}_\beta G_\mathrm{q}+\frac{\aone}{\sqrt{2\alpha_\mathrm{train}}}G_{\rm tr}
 +\sqrt{(\atwo-\aone^2)\Delta}\,G_\perp.
 \label{eq:scalar-prox}
\end{equation}
Define mean response $\chi(s)$ and susceptibility $s$ jointly by
\begin{equation}
 \chi(s)=\E\!\left[\frac1{1+s\sigma'(R_s)}\right], \quad \chi(s)-\frac{\lambda}{\Delta}s = 1 - \frac{1}{\tau}.
 \label{eq:s}
\end{equation}
Then for positive $s$, the scalars $\aone$ and $\atwo$ are desired solutions when
\begin{align}
 \aone=\frac{\E\!\left[R_s\left(\mathcal{C}_\beta G_\mathrm{q}+G_{\rm tr}/\sqrt{2\alpha_\mathrm{train}}\right)\right]}{\Delta + s\lambda}, \quad \atwo=\frac{\E[R_s^2]-\Delta(1-\chi)\aone^2 - s\lambda \aone^2}
       {\Delta\chi+s\lambda}.
\end{align}
\end{boxedlemmabrown}

These self-consistently determined scalars characterise the behaviour of the test logit $R=r_{\widehat{\Gamma}}(Z_\mathrm{new})$ of a fresh prompt $Z_\mathrm{new}$, as given by the following main result. Before proceeding, we need to be careful when discussing the fixed point solutions $a_1$ and $a_2$ for a given data configuration determined by $\alpha_\mathrm{train}, \tau, \beta$ and $\lambda$, as it is not clear if such \textit{finite} solutions exist. This is a well-known problem in logistic regression in the absence of regularisation \citep{candes2018phasetransitionexistencemaximum}.

Simply stated, if $n$ data points in $d$ dimensions have binary labels such that the two classes are linearly separable (\textit{i.e.}, by a hyperplane), then there is no finite maximum likelihood estimator: the likelihood can be increased by increasing the norm of the classifier weights without bound. If we assign independent random binary labels $0$ and $1$ with probability $1/2$ to $n$ $d$-dimensional points in general position, separability holds with probability tending to 1 as $n,d\to\infty$ if $n/d < 2$ \citep{cover1965geometrical,gardner1988space}, and thus MLE exists with high probability if $n/d > 2$. However, logistic regression makes it more difficult to determine the separability threshold as now the labels are no longer uniform, but themselves depend on the data through $\task^\top \x$. \cite{candes2018phasetransitionexistencemaximum} and later \cite{koriyama2025phasetransitionsexistenceunregularized} show that there is a relationship between the effective signal strength and the transition value of $n/d$. The zero-signal case corresponds to the Cover transition above where the critical value of $n/d$ is 2.

\begin{boxedlemmabrown}{Ridgeless solution existence boundary}\label{lemma:critical_tau}
A finite-norm optimiser $\widehat{\Gamma}$ exists with high probability for $\tau > \tau_c$, where $\tau_c$ is defined by \begin{equation}
 \tau_c^{-1}=\inf_{t\in\R}\E\!\left[\max(0,G+tU)^2(1-\sigma(\beta G_\mathrm{q}))
 +\min(0,G+tU)^2\sigma(\beta G_\mathrm{q})\right]
 \label{eq:threshold}
\end{equation}
for $G,G_\mathrm{q},G_{\rm tr}\iid\mathcal{N}(0,1)$ and $U=(\mathcal{C}_\beta G_\mathrm{q}+G_{\rm tr}/\sqrt{2\alpha_\mathrm{train}})/\sqrt\Delta.$
\end{boxedlemmabrown}
We expect a similar phenomenon to occur in the ridgeless case of our theory; we have formalised this intuition in Result \ref{lemma:critical_tau}; see Appendix \ref{sec:informal_solution_boundary_proof} for an informal argument. Here the correct order parameter instead is $n/(d(d+1)) \approx n/d^2 = \tau$, as our estimator takes $n$ samples ($n$ sequences) of features $\h$ in $d(d+1)$ dimensions. The {zero-signal} case would correspond to each of the $n$ query labels $y_\mathrm{q}$ being chosen uniformly at random, and then the MLE should exist for $\tau > 2$.  Increasing $\beta$, or increasing $\alpha_\mathrm{train}$ so that the context more accurately represents its latent task, should increase the transition value away from 2. Indeed, empirically, we find a very good match between the location of empirical error blowup and the above predicted $\tau_c$, as seen by Figure \ref{fig:main-cluster}. Note that $\tau_c$ depends on both $\beta$ and $\alpha_\mathrm{train}$, as will be discussed below (see Figure \ref{fig:critical_tau}).

\begin{figure}[htbp]
    \centering
    \preprintgraphic[width=\linewidth]{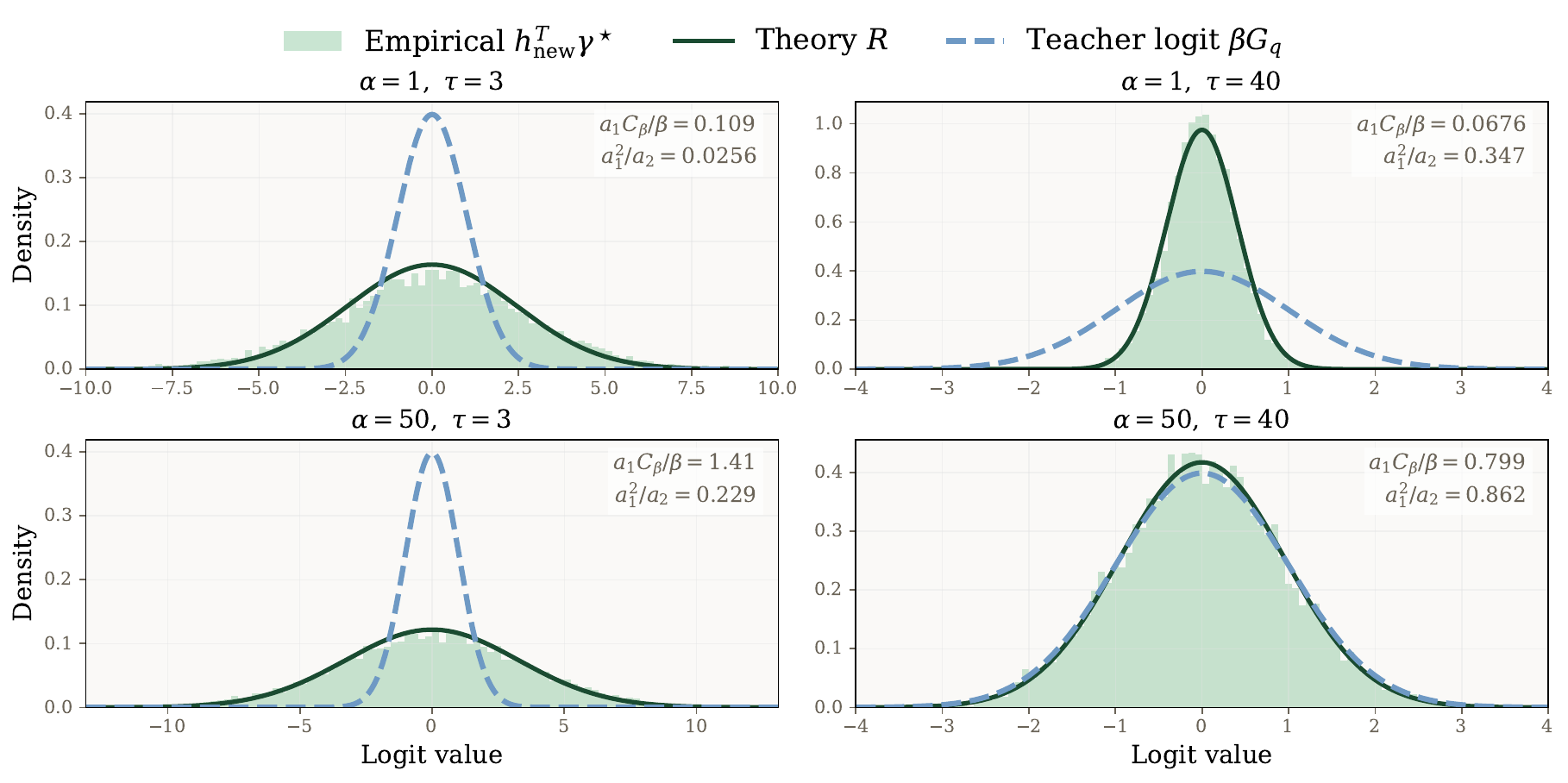}
    \caption{The distribution of empirical $\h_\text{test}^\top\bgamma^*$ matches our derived distribution $R$. The ridgeless theory predicts recovery of the true logit distribution as ${\alpha_\mathrm{train}},\tau\to\infty$. Increasing $\tau$ (right-hand plots) reduces variance, but alignment with $G_\mathrm{true}$ requires both variance reduction \textit{and} $a_1\mathcal{C}_\beta/\beta \to 1.$ For all, $\beta = 1, \lambda = 10^{-6}, d=32, \alpha_\mathrm{test}=\alpha_\mathrm{train}.$}
    \label{fig:spectrum_compare}
\end{figure}

\subsection{Finite-sample results: test logit and test error}
Finally, we may use the self-consistency equations in Result \ref{lemma:a1a2eqs} to characterise $\widehat{\bgamma}$, and to determine the ICL test error. To allow for different context length at testing time, we introduce $\alpha_\mathrm{test} = \ell_\mathrm{test}/d$.

\begin{boxedproposition}{Characterisation of logit distribution}
\label{prop:main_result_ridgeless}
Fix $\alpha_\mathrm{train},\tau,\beta,$ and $\lambda$. Sample Gaussians $G_\mathrm{q},G_\text{var}\sim_\mathrm{i.i.d.}\mathcal{N}(0,1)$. Then, assuming the empirical order parameters of the fitted $\widehat\Gamma$ concentrate to finite values, their limiting behaviour is
\begin{equation}
    (\hataone,\hatatwo,\hatazero) \approx (\aone,\atwo,0).
\end{equation}
Here $\aone$ and $\atwo$ are the deterministic self-consistent solutions given by Result \ref{lemma:a1a2eqs}. Further, the test logit $R=r_{\widehat\Gamma}(Z_{\rm new})$ has marginal distribution given by
\begin{align}
\text{true query target } y_\text{q}&\sim \text{Bernoulli}(\sigma(\beta G_\text{q} ))\label{eq:yquerymarginal} \\
\text{predicted query logit } R&=\aone\mathcal{C}_\beta G_\mathrm{q}+ G_\text{var}\sqrt{\frac{\atwo}{2\alpha_\mathrm{test}}+ (\atwo-\aone^2)\mathcal{C}_\beta^2}.
 \label{eq:rloo}
\end{align}
\end{boxedproposition}
The lack of $a_0$ in the final logit equation (i.e. $\widehat{a}_0 \approx 0$) is due to the isotropy and mean-0 properties of the data, as well as the fact that this corresponding block in the predictor, $\bm{a}^\top\x_\mathrm{q}/2$, contains no task information.

We provide empirical verification of this result in Figure \ref{fig:spectrum_compare} for a range of data parameters. Note that both the theoretical and empirical distributions tend towards the true logit distribution as $\alpha_\mathrm{train}$ and $\tau$ increase, as expected from the population analysis.

\begin{figure}[htb!]
    \centering
    \begin{subfigure}[b]{0.4\textwidth}
        \centering
        \preprintgraphic[width=\textwidth]{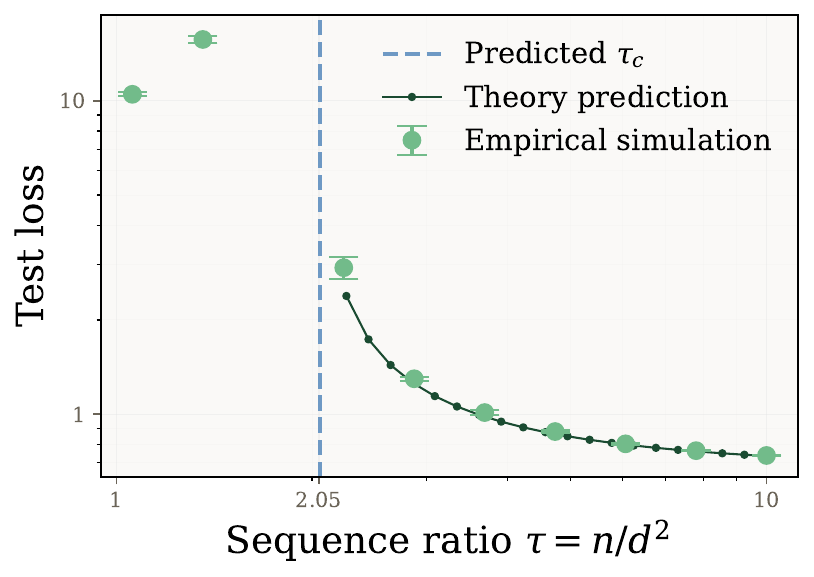}
        \caption{Ridgeless curve ($\lambda=0$)}
        \label{fig:sub-first}
    \end{subfigure}
    \hspace{0.5cm}
    \begin{subfigure}[b]{0.4\textwidth}
        \centering
        \preprintgraphic[width=\textwidth]{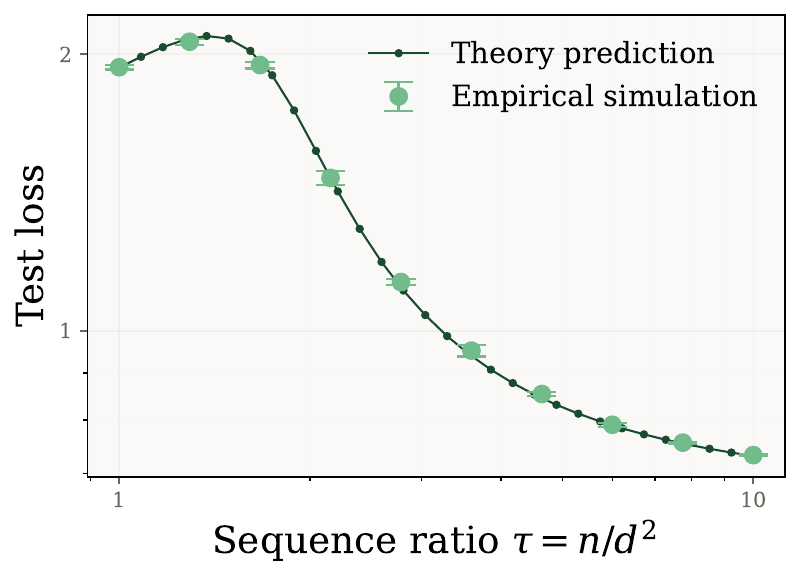}
        \caption{Non-zero ridge curve ($\lambda=0.001$)}
        \label{fig:sub-second}
    \end{subfigure}

    \caption{Empirical simulation refers to directly sampling $n$ sequences of length $\ell$, computing $\widehat{\bgamma}$, and evaluating this parameter matrix on freshly sampled sequences. This is compared to the theoretical prediction in Result \ref{prop:errors}. The critical ratio is given by eq. (\ref{eq:threshold}).  Here $d=64$, $\alpha_\mathrm{train}=4=\alpha_\mathrm{test}$, $\beta=1$. In (a), the empirical simulations take $\lambda = 10^{-8}$.}
    \label{fig:main-cluster}
\end{figure}

\begin{figure}[htb!]
    \centering
    \preprintgraphic[width=0.4\linewidth]{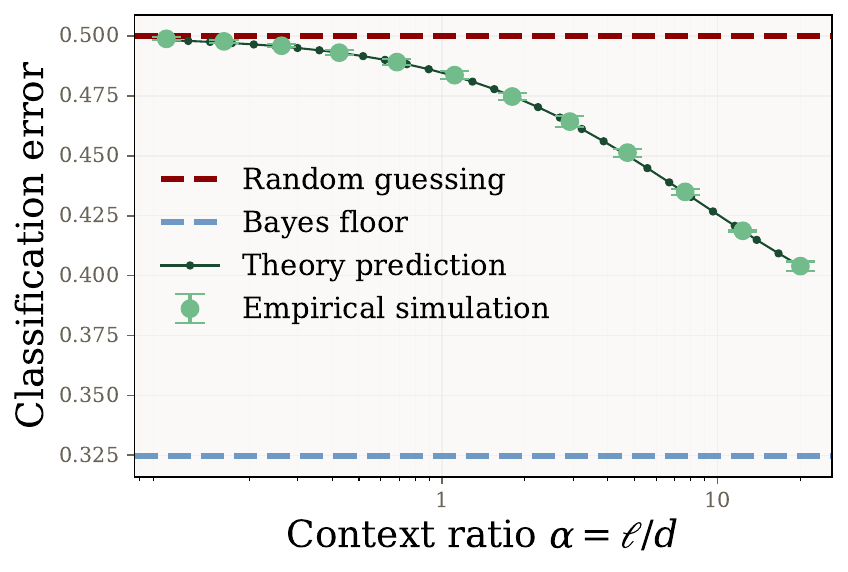}
    \caption{Simulation versus theory for classification error. We can compare to random guessing error 1/2, and Bayes error $1-\E[\sigma(\beta\abs{G_\mathrm{q}})]$. For {short contexts}, the model effectively guesses, but nears the Bayes floor as context length increases. $d=32, \alpha_\mathrm{train} = \alpha_\mathrm{test}, \tau = 10, \beta = 1, \lambda = 10^{-8}$.}
    \label{fig:class_error_alpha}
\end{figure}

This characterisation of the joint distribution of the latent and predicted logits makes it easy to compute the limiting cross-entropy loss and classification error, verified in Figures \ref{fig:main-cluster} and \ref{fig:class_error_alpha}.

\begin{boxedproposition}{Error formulas}\label{prop:errors}
The cross entropy test loss is given by \begin{align}
    \mathcal{E}_\text{CE} &= \E_{G_\mathrm{q},G_\text{var}}\big[L(R)-\sigma(\beta G_\mathrm{q})R\big]
\label{eq:test_loss_final_expression}
\end{align} and the classification error is given by \begin{align}
    \mathcal{E}_\text{class} = \mathbb{P}\!\left(\bm{1}\{R\geq0\}\neq Y\right) = \E_{G_\mathrm{q}}\qty[\sigma(\beta G_\mathrm{q})\qty(1-\phi)+\qty(1-\sigma(\beta G_\mathrm{q}))\phi]
\end{align} where $\Phi$ is the standard normal CDF and $\phi = \Phi\qty(a_1\mathcal{C}_\beta G_\mathrm{q}/{\sqrt{V}}),\;\; V = {a_2}/({2\alpha_\mathrm{test}}) +(a_2-a_1^2)\mathcal{C}_\beta^2.$
\end{boxedproposition}

\section{Findings}
We highlight several interpretable implications of these results.

\paragraph{Effect of test-time context length} First note that $\alpha_\text{test}$ only appears explicitly in $V$; $a_1$ and $a_2$ do not depend on $\alpha_\mathrm{test}$. As $\alpha_\text{test}$ increases, this contribution to non-signal variance will vanish and ICL error will strictly decrease, both as measured by cross-entropy loss or by classification error (see Appendix \ref{sec:test_time_monotonic} for proof). This is an artifact of our choice of linear attention architecture, which explicitly averages its inputs through the $1/\ell$ term that appears in the architecture definition (\ref{eq:linear-attention}); thus, longer test contexts always reduce noise and improve error \citep{letey2025pretrain}. Whether or not this particular phenomenon generalises to other architectures is a question for future work.


\paragraph{Calibration and noise} Under the marginal laws (\ref{eq:yquerymarginal}) and (\ref{eq:rloo}), the target logit is $G_{\rm true}=\beta G_\mathrm{q}$ and the predicted logit is
\begin{equation}
 R=\frac{\aone\mathcal{C}_\beta}{\beta}G_{\rm true}+\sqrt{V}\,G_\text{var},\quad
 V=\frac{\atwo}{2\alpha_\text{test}} + (\atwo - \aone^2) \, \mathcal{C}_\beta^2,\quad G_\text{var}\sim\cN(0,1). \nonumber
\end{equation}

By Result~\ref{prop:main_result_ridgeless}, this logit expression makes the various sources of error explicit: the scaling coefficient $a_1\mathcal{C}_\beta/\beta$ in front of $G_\mathrm{true}$, the variance due to test context length $\alpha_\text{test}$, and the variance due to anisotropy $a_2-a_1^2$. While the factor $a_1\mathcal{C}_\beta/\beta$ might initially appear to be a direct measure of ``calibration'' or alignment between our prediction and the true label, we must realise that each of these factors interact nontrivially. To see this, write the conditional distribution of the target in terms of the prediction, namely,

$$G_\text{true}\mid R \sim \mathcal{N}\qty(\frac{a_1\beta \mathcal{C}_\beta}{a_1^2 \mathcal{C}_\beta^2 + V}R,\,\, \frac{\beta^2}{a_1^2 \mathcal{C}_\beta^2 + V}V).$$

This immediately allows us to write the conditional class probability as \begin{equation}\label{eq:logit_likelihood}
    p(R):=\mathbb{P}(Y=1\mid R) = \E_{Z\sim \mathcal{N}(0,1)}\qty[\sigma(mR + vZ)]\,,\quad m = \frac{a_1\beta \mathcal{C}_\beta}{a_1^2\mathcal{C}^2_\beta + V}\,, \quad v^2 = \frac{\beta^2}{a_1^2\mathcal{C}^2_\beta + V}V.
\end{equation}
Recall the model predicts the probability of outcome 1 as $\sigma(R)$, where the actual probability is $p(R)$. This immediately gives us a clean formulation of calibration: the model is \textbf{calibrated} when $p(R)=\sigma(R)$. For $R > 0$ (i.e. the model is skewed towards 1), if $\sigma(R) < p(R)$ then the model is underconfident, while if $\sigma(R) > p(R)$ the model is overconfident. The reverse is true for $R < 0$.

\begin{figure}[htbp]
    \centering
    \preprintgraphic[width=\linewidth]{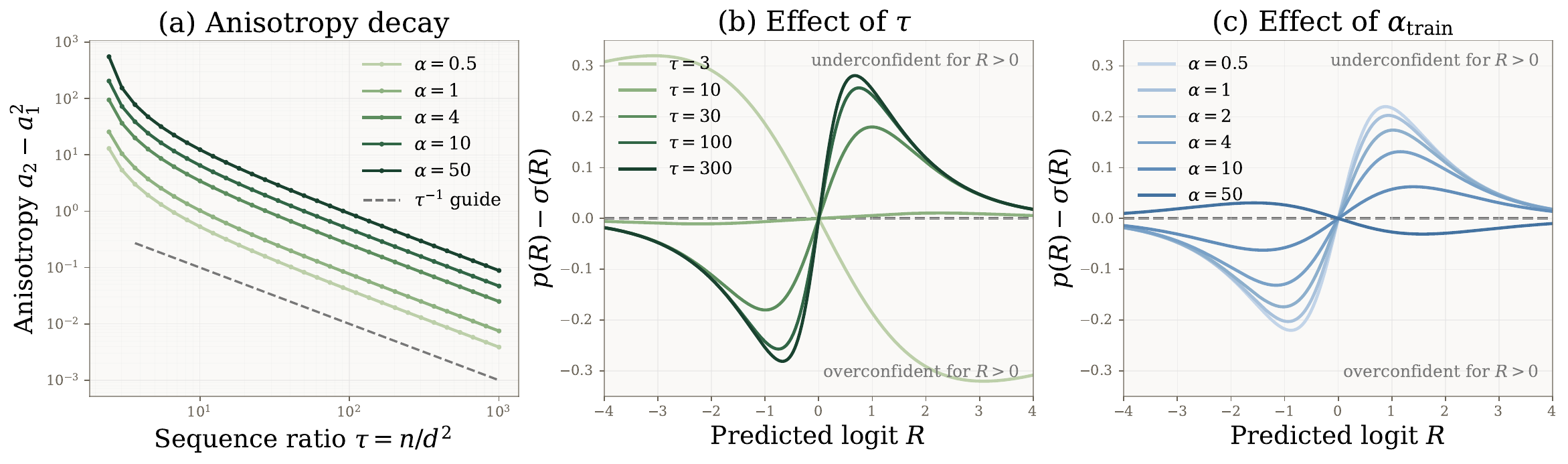}
    \caption{Effects of data parameters $\alpha_\mathrm{train}$ and $\tau$ on anisotropy (a) and calibration (b),(c). We compute $a_1, a_2$ (and correspondingly, $m,v$) using Result \ref{lemma:a1a2eqs}. In (b),(c), $R$ is taken simply as a real-valued variable and the function $p(R)$ is computed by numerically taking the expectation in (\ref{eq:logit_likelihood}). In all plots we take $\alpha_\mathrm{test}=10^8$ (effectively infinite) to isolate the effects of $\alpha_\mathrm{train}$ and $\tau$; all labels $\alpha$ refer to $\alpha_\mathrm{train}$. For all, $\lambda=0$ and $\beta = 1$. For (b) $\alpha_\mathrm{train} = 4$. For (c) $\tau = 20$.}
    \label{fig:calibration}
\end{figure}

Figure \ref{fig:calibration} demonstrates the effect of data finiteness, on both anisotropy and calibration. We indeed recover the population intuition that additional pretraining data induces isotropy, as $a_2-a_1^2 \to 0$ as $\tau \to \infty$. We see this occurs roughly at a rate of $\tau^{-1}$, although this remains unproven. Additionally, we see the nontrivial interactions of context length and pretraining sequences on calibration in Figures \ref{fig:calibration}(b) and \ref{fig:calibration}(c). In particular, while increasing $\tau$ reduces error due to variance, it does not necessarily produce better calibration. Depending on the data configuration, the model may instead move between underconfidence and overconfidence. These nontrivial nonmonotonicities deserve further exploration, which we leave to future work.

\begin{wrapfigure}{r}{0.5\textwidth}
    \centering
    \preprintgraphic[width=0.4\textwidth]{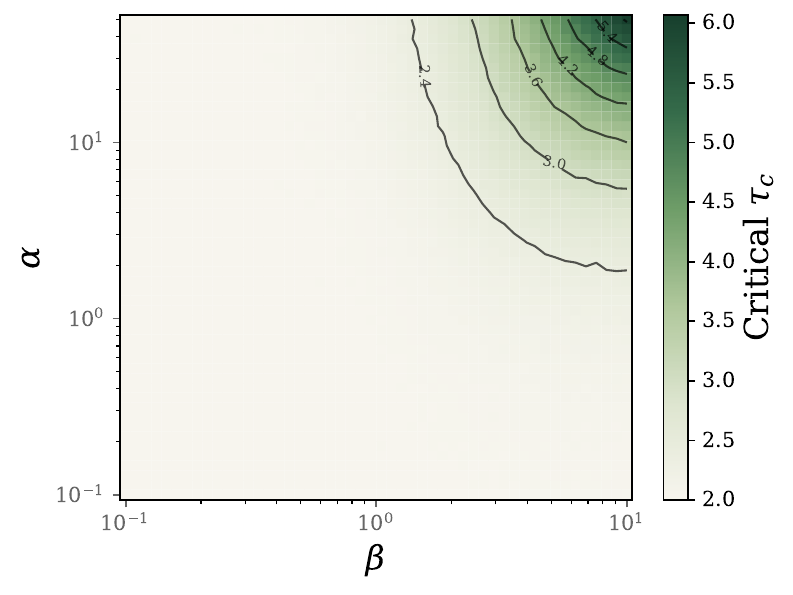}
    \caption{Heatmap of critical $\tau_c$ against $\alpha$ and $\beta$.}
    \label{fig:critical_tau}
\end{wrapfigure}

\paragraph{Location of critical batch size} Figure \ref{fig:critical_tau} maps the predicted dependence of $\tau_c$ on context length $\alpha_\mathrm{train}$ and sampling inverse temperature $\beta$. We see that increasing either $\alpha_\mathrm{train}$ or $\beta$ will increase the critical ratio $\tau_c$. This can be understood from prior intuition:  increasing either $\alpha_\mathrm{train}$ or $\beta$ will increase the "signal" of the features relative to the zero-signal (uniformly random $y_\text{q}$s) case, thus making separability easier to achieve, and pushing the critical sample ratio away from 2.

\section{Beyond the reduced model}
\begin{figure}[htbp]
    \centering
    \preprintgraphic[width=\linewidth]{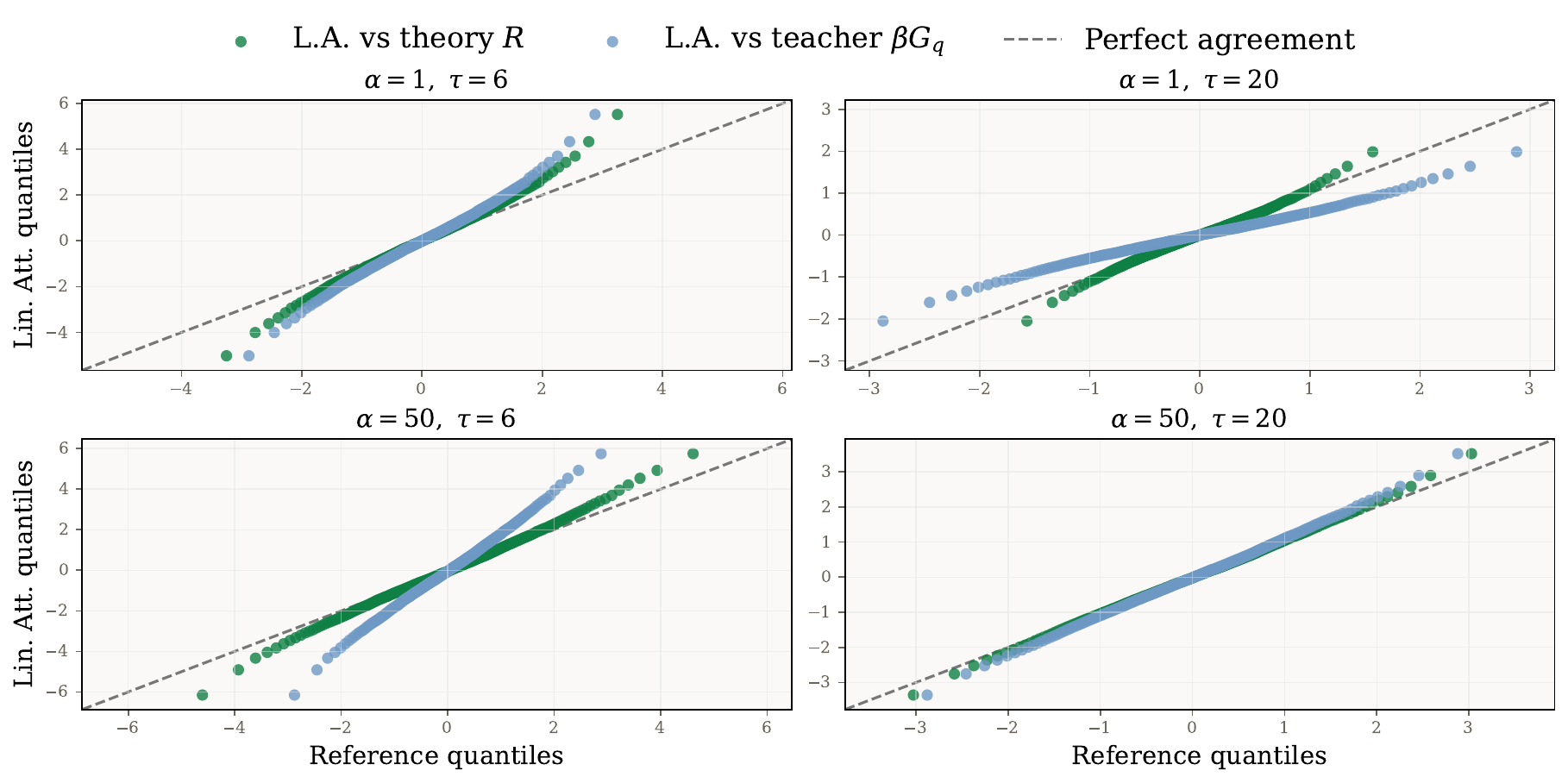}
    \caption{Similarly to Figure \ref{fig:spectrum_compare}, we compare the theory to full-parameter linear attention, namely, we train the full $K,Q,V$ matrices. Here we plot a quantile-quantile comparison of the model's output logit on new prompt matrices $Z$ against the theory prediction $R$ and the true latent $\beta G_\mathrm{q}$. In all plots, $d=16, \lambda = 10^{-6}, \beta=1$. Regularisation is applied to $\|K\|_F^2 + \|Q\|_F^2 + \|V\|_F^2$. }
    \label{fig:spectrum_LA}
\end{figure}

It is important to characterise how well this model extends beyond the theoretical regime. Recall we began with the fully-parameterised linear attention architecture in terms of $K,Q,V$ matrices, and ultimately studied the reduced model in terms of parameters $\Gamma$. Despite this modelling choice for the theory, the prediction in Result \ref{prop:main_result_ridgeless} also accurately describes the linear-attention architecture with fully trained $K, Q, V$ matrices. This is shown in Figure \ref{fig:spectrum_LA}, where we see close agreement between the theory logit $R$ and the trained linear attention output distributions, with small deviation in the tails for small $\alpha$ and $\tau$.

\begin{figure}[htbp]
    \centering
    \preprintgraphic[width=\linewidth]{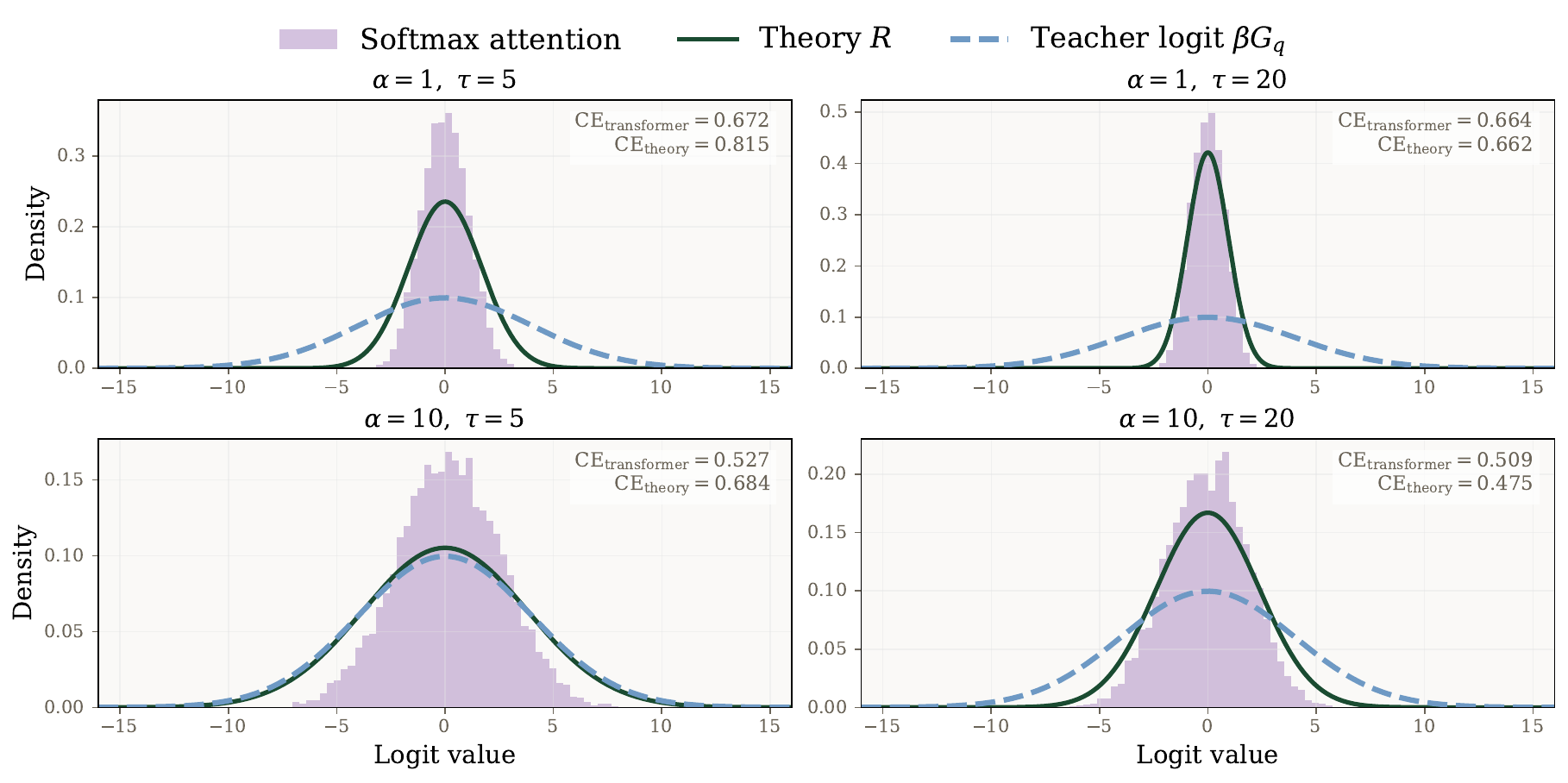}
    \caption{An analogous comparison to Figure \ref{fig:spectrum_compare}, but the logits here are from a trained transformer-style softmax attention block. In particular, we use standard scaled dot-product softmax attention with fully trainable query $Q$, key $K$, value $V$, and output $O$ weights. We initialise these at identity to improve training dynamics. The model is trained for 100 epochs with AdamW with weight decay 0.1 and early stopping. We set $d=32$ and $\beta = 4$.}
    \label{fig:softmax}
\end{figure}

We do not expect the exact theoretical characterisation to transfer directly to other architectures. In particular, conclusions about test-time context length likely depend on the explicit averaging structure of linear attention. Regardless, Figure \ref{fig:softmax} shows that the trained softmax-attention model exhibits some qualitatively similar behaviour. We include the linear-attention theory distribution strictly as a reference; it is not a theoretical prediction for the softmax model.

Across all four settings, the softmax logits have noticeably lower variance than both the linear-attention theory prediction and the teacher logits. Increasing the context ratio $\alpha$ substantially broadens the softmax distribution and improves its alignment with the teacher distribution. The effect of increasing the task ratio $\tau$ is less straightforward: it reduces the softmax-logit variance in these experiments rather than producing uniformly closer agreement with the teacher. The softmax model achieves lower cross-entropy than the linear-attention theory reference for $\tau=5$, whereas the theory reference performs better for $\tau=20$. These results suggest potentially interesting differences in calibration and confidence between the two architectures, which we leave for future investigation.

\section{Conclusion}
We have characterised the high-dimensional asymptotics of in-context binary classification, using a reduced linear-attention model with a softmax readout. The resulting joint law of the latent and predicted logits separates three distinct effects of data on ICL performance: scaling of the teacher-aligned signal, noise from finite test contexts, and anisotropy induced by finitely many pretraining tasks. This decomposition yields explicit predictions for cross-entropy loss, classification error. Furthermore, we identify the separability boundary beyond which the ridgeless logistic estimator has no finite solution. For all our results, finite-dimensional simulations closely match these predictions, and fully trained linear-attention models resemble closely our predicted logit distribution.

Our analysis also shows that improvements in loss and calibration need not match. Longer test contexts monotonically remove a source of noise in the reduced architecture, whereas increasing the number of pretraining tasks removes anisotropic noise without necessarily improving calibration. Consequently, calibration may vary nonmonotonically even as predictive error improves. Our theory and its predictions are specific to the reduced linear-attention setting. We provide softmax-attention experiments as qualitative evidence beyond that regime. Extending the calculation to richer architectures is a natural direction for future work.

\section*{Acknowledgments and Disclosure of Funding}
We thank Luca Pesce for helpful discussions. A.R. was supported by the Alumni Sponsored Internship Program (ASIP) of Williams College. M.L. is supported by a Graduate Fellowship from the Kempner Institute for the Study of Natural and Artificial Intelligence. Y.M.L. gratefully acknowledges support from a Harvard College Professorship, the Harvard FAS Dean’s Fund for Promising Scholarship, and DARPA grant DIAL-FP-038. J.Z.-V. was supported by a Junior Fellowship from the Harvard Society of Fellows. C.P. is supported by an NSF CAREER Award (IIS-2239780), DARPA grants DIAL-FP-038 and AIQ-HR00112520041, and the Simons Collaboration on the Physics of Learning and Neural Computation. This work has been made possible in part by a gift from the Chan Zuckerberg Initiative Foundation to establish the Kempner Institute for the Study of Natural and Artificial Intelligence.

\section*{AI use statement}
We used generative AI tools for various steps throughout the research and writing processes. We describe this as follows. 
\begin{itemize}
    \item \textcolor{gray}{Synthetic dataset generation.} \textcolor{black}{No.} 
    \item \textcolor{gray}{Develop theoretical models.} \textcolor{black}{No.} 
    \item \textcolor{gray}{Formulate mathematical claims.} \textcolor{black}{AI was used to help state our formulas more clearly in the document; it was not used to formulate the actual mathematics.} 
    \item \textcolor{gray}{Provide critical components for proofs.} \textcolor{black}{No.} 
    \item \textcolor{gray}{Assist in the writing of proofs.} \textcolor{black}{Yes. GPT 5.6 and 6 were used to verify and check our proofs and formulas; GPT 6 suggested a simplification of an expression in Appendix~\ref{app:derivation} such that (\ref{eq:s}) does not involve the inverse Hessian trace explicitly, which helps with numerical stability.}
    \item \textcolor{gray}{Propose or refine hypotheses.} \textcolor{black}{No.}
    \item \textcolor{gray}{Design research methodology.} \textcolor{black}{No.}
    \item \textcolor{gray}{Provide feedback on research methodology.} \textcolor{black}{Yes, namely verification of our proof and method.} 
    \item \textcolor{gray}{Implement methods.} \textcolor{black}{Yes. GPT 5.6 and GPT 6 were used to generate substantial blocks of the code, particularly for simulations of the theory; all implementations and code were commented, rewritten, and verified by the lead authors. The most significant use of AI to generate code was for the final softmax experiments, which we offer as a preliminary investigation.}
    \item \textcolor{gray}{Assist with translation.} \textcolor{black}{No.}
    \item \textcolor{gray}{Clean and reformat datasets.} \textcolor{black}{No.}
    \item \textcolor{gray}{Support data analysis.} \textcolor{black}{No.}
    \item \textcolor{gray}{Interpret results.} \textcolor{black}{AI was used in the writing process, with specific prompts, where we asked specific questions in order to improve the clarity, readability, and logical flow of our results. Some of this naturally includes interpretation. Additionally, upon a review pass of the manuscript, GPT 5.6 suggested the comparison of $p(R)$ (given by equation (\ref{eq:logit_likelihood})) with $\sigma(R)$. All additional interpretation and experiment design were then done by the authors.}
    \item \textcolor{gray}{Brainstorming.} \textcolor{black}{Yes. The conception of this paper occurred through a brainstorming session between the senior author and an agent model. In particular, this included the suggestion that the softmax-readout linear attention architecture might be analysable at the population level, and suggested the initial rotational symmetry approach.} 
\end{itemize}

AI was not used to draft the manuscript or generate the literature review; it was used to polish prose and for suggestions on overall presentation. We have reviewed all AI-assisted work. We take responsibility for the final content of this work,
including text, claims, or artefacts produced with the aid of generative AI.

\clearpage
\IfFileExists{references.bib}{%
  \bibliography{references}

@inproceedings{akyurek2023what,
	title        = {What learning algorithm is in-context learning? Investigations with linear models},
	author       = {Ekin Aky{\"u}rek and Dale Schuurmans and Jacob Andreas and Tengyu Ma and Denny Zhou},
	year         = {2023},
	booktitle    = {The Eleventh International Conference on Learning Representations},
	url          = {https://openreview.net/forum?id=0g0X4H8yN4I}
}

@article{barbier2019optimal,
  title={Optimal errors and phase transitions in high-dimensional generalized linear models},
  author={Barbier, Jean and Krzakala, Florent and Macris, Nicolas and Miolane, L{\'e}o and Zdeborov{\'a}, Lenka},
  journal={Proceedings of the National Academy of Sciences},
  volume={116},
  number={12},
  pages={5451--5460},
  year={2019},
  publisher={National Academy of Sciences}
}

@inproceedings{bai2023transformers,
	title        = {Transformers as Statisticians: Provable In-Context Learning with In-Context Algorithm Selection},
	author       = {Bai, Yu and Chen, Fan and Wang, Huan and Xiong, Caiming and Mei, Song},
	year         = {2023},
	booktitle    = {Advances in Neural Information Processing Systems},
	publisher    = {Curran Associates, Inc.},
	volume       = {36},
	pages        = {57125--57211},
	url          = {https://proceedings.neurips.cc/paper_files/paper/2023/file/b2e63e36c57e153b9015fece2352a9f9-Paper-Conference.pdf},
	editor       = {A. Oh and T. Naumann and A. Globerson and K. Saenko and M. Hardt and S. Levine}
}

@misc{bordelon2025theoryscalinglawsincontext,
	title        = {Theory of Scaling Laws for In-Context Regression: Depth, Width, Context and Time},
	author       = {Blake Bordelon and Mary I. Letey and Cengiz Pehlevan},
	year         = {2025},
	url          = {https://arxiv.org/abs/2510.01098},
	eprint       = {2510.01098},
	archiveprefix = {arXiv},
	primaryclass = {stat.ML}
}

@article{brown2020language,
	title        = {Language models are few-shot learners},
	author       = {Brown, Tom and Mann, Benjamin and Ryder, Nick and Subbiah, Melanie and Kaplan, Jared D and Dhariwal, Prafulla and Neelakantan, Arvind and Shyam, Pranav and Sastry, Girish and Askell, Amanda and others},
	year         = {2020},
	journal      = {Advances in neural information processing systems},
	volume       = {33},
	pages        = {1877--1901}
}

@misc{chan2022data,
	title        = {Data Distributional Properties Drive Emergent In-Context Learning in Transformers},
	author       = {Stephanie C. Y. Chan and Adam Santoro and Andrew K. Lampinen and Jane X. Wang and Aaditya Singh and Pierre H. Richemond and Jay McClelland and Felix Hill},
	year         = {2022},
	eprint       = {2205.05055},
	archiveprefix = {arXiv},
	primaryclass = {cs.LG}
}

@article{cover1965geometrical,
  title={Geometrical and statistical properties of systems of linear inequalities with applications in pattern recognition},
  author={Cover, Thomas M},
  journal={IEEE transactions on electronic computers},
  number={3},
  pages={326--334},
  year={1965},
  publisher={IEEE}
}

@article{zhang2026transformers,
  title={Transformers Efficiently Perform In-Context Logistic Regression via Normalized Gradient Descent},
  author={Zhang, Chenyang and Cao, Yuan},
  journal={arXiv preprint arXiv:2605.06609},
  year={2026}
}

@inproceedings{ahn2024linear,
  title={Linear attention is (maybe) all you need (to understand transformer optimization)},
  author={Ahn, Kwangjun and Cheng, Xiang and Song, Minhak and Yun, Chulhee and Jadbabaie, Ali and Sra, Suvrit},
  booktitle={International Conference on Learning Representations},
  volume={2024},
  pages={17686--17698},
  year={2024}
}

@article{gardner1988space,
doi = {10.1088/0305-4470/21/1/030},
url = {https://doi.org/10.1088/0305-4470/21/1/030},
year = {1988},
month = {jan},
publisher = {},
volume = {21},
number = {1},
pages = {257},
author = {Gardner, Elizabeth},
title = {The space of interactions in neural network models},
journal = {Journal of Physics A: Mathematical and General},
}

@INPROCEEDINGS{mai_largescale_logistic_asymp,
  author={Mai, Xiaoyi and Liao, Zhenyu and Couillet, Romain},
  booktitle={ICASSP 2019 - 2019 IEEE International Conference on Acoustics, Speech and Signal Processing (ICASSP)}, 
  title={A Large Scale Analysis of Logistic Regression: Asymptotic Performance and New Insights}, 
  year={2019},
  volume={},
  number={},
  pages={3357-3361},
  doi={10.1109/ICASSP.2019.8683376}}

@inproceedings{gerace2020generalisation,
	title        = {Generalisation error in learning with random features and the hidden manifold model},
	author       = {Gerace, Federica and Loureiro, Bruno and Krzakala, Florent and M{\'e}zard, Marc and Zdeborov{\'a}, Lenka},
	year         = {2020},
	booktitle    = {International Conference on Machine Learning},
	pages        = {3452--3462},
	organization = {PMLR}
}

@inproceedings{katharopoulos2020transformers,
	title        = {Transformers are {RNN}s: Fast autoregressive transformers with linear attention},
	author       = {Katharopoulos, Angelos and Vyas, Apoorv and Pappas, Nikolaos and Fleuret, Fran{\c{c}}ois},
	year         = {2020},
	booktitle    = {International conference on machine learning},
	pages        = {5156--5165},
	organization = {PMLR}
}

@inproceedings{letey2025pretrain,
	title        = {Pretrain-Test Task Alignment Governs Generalization in In-Context Learning},
	author       = {Mary Letey and Jacob A Zavatone-Veth and Yue M. Lu and Cengiz Pehlevan},
	year         = {2026},
	booktitle    = {The Fourteenth International Conference on Learning Representations},
	url          = {https://openreview.net/forum?id=KZLeg0MQ2r}
}

@inproceedings{letey2026sequential,
	title        = {Sequential Correlations Change In-Context Learning: Effective Context Length and Architectural Mismatch},
	author       = {Mary Letey and Yue M. Lu and Cengiz Pehlevan and Jacob A Zavatone-Veth},
	year         = {2026},
	booktitle    = {High-dimensional Learning Dynamics 2026},
	url          = {https://openreview.net/forum?id=AYprbQxAbI}
}

@article{lu2025asymptotictheoryincontextlearning,
	title        = {Asymptotic theory of in-context learning by linear attention},
	author       = {Yue M. Lu  and Mary Letey  and Jacob A. Zavatone-Veth  and Anindita Maiti  and Cengiz Pehlevan},
	year         = {2025},
	journal      = {Proceedings of the National Academy of Sciences},
	volume       = {122},
	number       = {28},
	pages        = {e2502599122},
	doi          = {10.1073/pnas.2502599122},
	url          = {https://www.pnas.org/doi/abs/10.1073/pnas.2502599122},
	eprint       = {https://www.pnas.org/doi/pdf/10.1073/pnas.2502599122}
}

@article{mei2022generalization,
	title        = {The generalization error of random features regression: Precise asymptotics and the double descent curve},
	author       = {Mei, Song and Montanari, Andrea},
	year         = {2022},
	journal      = {Communications on Pure and Applied Mathematics},
	publisher    = {Wiley Online Library},
	volume       = {75},
	number       = {4},
	pages        = {667--766}
}

@misc{oko2024pretrainedtransformerefficientlylearns,
	title        = {Pretrained transformer efficiently learns low-dimensional target functions in-context},
	author       = {Kazusato Oko and Yujin Song and Taiji Suzuki and Denny Wu},
	year         = {2024},
	url          = {https://arxiv.org/abs/2411.02544},
	eprint       = {2411.02544},
	archiveprefix = {arXiv},
	primaryclass = {cs.LG}
}

@misc{pesce2023gaussiandataneedextents,
	title        = {Are Gaussian data all you need? Extents and limits of universality in high-dimensional generalized linear estimation},
	author       = {Luca Pesce and Florent Krzakala and Bruno Loureiro and Ludovic Stephan},
	year         = {2023},
	url          = {https://arxiv.org/abs/2302.08923},
	eprint       = {2302.08923},
	archiveprefix = {arXiv},
	primaryclass = {math.ST}
}

@inproceedings{vasudeva2025transformers,
	title        = {Transformers learn low sensitivity functions: Investigations and implications},
	author       = {Vasudeva, Bhavya and Fu, Deqing and Zhou, Tianyi and Kau, Elliott and Huang, Youqi and Sharan, Vatsal},
	year         = {2025},
	booktitle    = {International Conference on Learning Representations},
	volume       = {2025},
	pages        = {7805--7834}
}

@inproceedings{vonoswald2023transformers,
	title        = {Transformers Learn In-Context by Gradient Descent},
	author       = {Von Oswald, Johannes and Niklasson, Eyvind and Randazzo, Ettore and Sacramento, João and Mordvintsev, Alexander and Zhmoginov, Andrey and Vladymyrov, Max},
	year         = {2023},
	month        = {23--29 Jul},
	booktitle    = {Proceedings of the 40th International Conference on Machine Learning},
	publisher    = {PMLR},
	series       = {Proceedings of Machine Learning Research},
	volume       = {202},
	pages        = {35151--35174},
	url          = {https://proceedings.mlr.press/v202/von-oswald23a.html},
	editor       = {Krause, Andreas and Brunskill, Emma and Cho, Kyunghyun and Engelhardt, Barbara and Sabato, Sivan and Scarlett, Jonathan}
}

@article{wei2022emergent,
	title        = {Emergent Abilities of Large Language Models},
	author       = {Jason Wei and Yi Tay and Rishi Bommasani and Colin Raffel and Barret Zoph and Sebastian Borgeaud and Dani Yogatama and Maarten Bosma and Denny Zhou and Donald Metzler and Ed H. Chi and Tatsunori Hashimoto and Oriol Vinyals and Percy Liang and Jeff Dean and William Fedus},
	year         = {2022},
	journal      = {Transactions on Machine Learning Research},
	issn         = {2835-8856},
	url          = {https://openreview.net/forum?id=yzkSU5zdwD}
}

@inproceedings{wu2023pretraining,
	title        = {How Many Pretraining Tasks Are Needed for In-Context Learning of Linear Regression?},
	author       = {Jingfeng Wu and Difan Zou and Zixiang Chen and Vladimir Braverman and Quanquan Gu and Peter Bartlett},
	year         = {2024},
	booktitle    = {The Twelfth International Conference on Learning Representations},
	url          = {https://openreview.net/forum?id=vSh5ePa0ph}
}

@misc{zhang2023doesincontextlearninglearn,
	title        = {What and How does In-Context Learning Learn? Bayesian Model Averaging, Parameterization, and Generalization},
	author       = {Yufeng Zhang and Fengzhuo Zhang and Zhuoran Yang and Zhaoran Wang},
	year         = {2023},
	url          = {https://arxiv.org/abs/2305.19420},
	eprint       = {2305.19420},
	archiveprefix = {arXiv},
	primaryclass = {stat.ML}
}

@article{zhang2023trained,
	title        = {Trained Transformers Learn Linear Models In-Context},
	author       = {Ruiqi Zhang and Spencer Frei and Peter L. Bartlett},
	year         = {2024},
	journal      = {Journal of Machine Learning Research},
	volume       = {25},
	number       = {49},
	pages        = {1--55},
	url          = {http://jmlr.org/papers/v25/23-1042.html}
}

@misc{candes2018phasetransitionexistencemaximum,
      title={The phase transition for the existence of the maximum likelihood estimate in high-dimensional logistic regression}, 
      author={Emmanuel J. Candes and Pragya Sur},
      year={2018},
      eprint={1804.09753},
      archivePrefix={arXiv},
      primaryClass={stat.ME},
      url={https://arxiv.org/abs/1804.09753}, 
}

@article{Sur_2019,
   title={A modern maximum-likelihood theory for high-dimensional logistic regression},
   volume={116},
   ISSN={1091-6490},
   url={http://dx.doi.org/10.1073/pnas.1810420116},
   DOI={10.1073/pnas.1810420116},
   number={29},
   journal={Proceedings of the National Academy of Sciences},
   publisher={National Academy of Sciences},
   author={Sur, Pragya and Candès, Emmanuel J.},
   year={2019},
   month=July, pages={14516–14525} }

@article{el2018impact,
  title={On the impact of predictor geometry on the performance on high-dimensional ridge-regularized generalized robust regression estimators},
  author={El Karoui, Noureddine},
  journal={Probability Theory and Related Fields},
  volume={170},
  number={1},
  pages={95--175},
  year={2018},
  publisher={Springer}
}

@misc{koriyama2025phasetransitionsexistenceunregularized,
      title={Phase transitions for the existence of unregularized M-estimators in single index models} , 
      author={Takuya Koriyama and Pierre C. Bellec},
      year={2025},
      eprint={2501.03163},
      archivePrefix={arXiv},
      primaryClass={math.ST},
      url={https://arxiv.org/abs/2501.03163}, 
}
  \bibliographystyle{plainnat}
}{%
  \begin{center}\small\color{PreprintMuted}
  Bibliography omitted in this single-file preview. The project build uses
  \texttt{references.bib}.
  \end{center}
}

\newpage
\appendix
\section{Numerics and code}
Code and additional numerical details will be made available on \href{https://github.com/mletey/in-context-classification-with-linear-attention}{GitHub}.

\section{Derivation of the cavity result}
\label{app:derivation}
The following provides a derivation of Result \ref{lemma:a1a2eqs} and Result \ref{prop:main_result_ridgeless} together. Note that Result \ref{prop:errors} follows immediately from \ref{prop:main_result_ridgeless}.

\begin{proof} We will decompose the proof into steps.

\paragraph{Step 1. Proximal equation} Write $f_y(z)=L(z)-yz$ and let $\widehat\bgamma_{[\nu]}$ minimise the objective with prompt $\nu$ removed. Let
\begin{equation}
 \mathcal H_{[\nu]}=\frac1n\sum_{\mu\ne\nu}\sigma'(\fitlogit{\mu,[\nu]})\bh_\mu\bh_\mu^\top+ \lambda d  I_{d(d+1)},
 \quad \fitlogit{\mu,[\nu]}=\bh_\mu^\top\widehat\bgamma_{[\nu]}.
\end{equation}
A quadratic expansion about $\widehat\bgamma_{[\nu]}$ gives
\begin{equation}
 \widehat\bgamma\approx\arg\min_{\bgamma}\left\{
 \frac1n f_{y_\nu}(\bh_\nu^\top\bgamma)
 +\frac12(\bgamma-\widehat\bgamma_{[\nu]})^\top\mathcal H_{[\nu]}
 (\bgamma-\widehat\bgamma_{[\nu]})\right\}.
\end{equation}
Minimizing at fixed $z=\bh_\nu^\top\bgamma$ yields
\begin{equation}
 \fitlogit{\nu}=\prox_{s_\nu f_{y_\nu}}(\fitlogit{\nu,[\nu]}),
 \qquad
 \fitlogit{\nu,[\nu]}=\fitlogit{\nu}+s_\nu(\sigma(\fitlogit{\nu})-y_\nu),
 \qquad
 s_\nu=\frac1n\bh_\nu^\top\mathcal H_{[\nu]}^{-1}\bh_\nu.
 \label{eq:appendix-prox}
\end{equation}
\textbf{Logit convention.} Before taking the high-dimensional limit, $\fitlogit{\mu}:=\bh_\mu^\top\widehat\bgamma$ denotes the full-sample fitted logit, while $\fitlogit{\mu,[\nu]}:=\bh_\mu^\top\widehat\bgamma_{[\nu]}$ denotes the logit of prompt $\mu$ when prompt $\nu$ is omitted from training. In particular, $\fitlogit{\mu,[\mu]}$ is the cavity logit for the held-out prompt. Only after invoking concentration and exchangeability do we identify the limiting joint law of $(\fitlogit{\mu,[\mu]},\fitlogit{\mu},y_\mu)$ with $(R_{\mathrm{loo}},R_s,Y)$.

\paragraph{Step 2. Hessian prefactor} We are interested in writing a characterisation for the left-out Hessian prefactor terms that appear in the proximal equation
$$
s_\nu := \frac{1}{n}\bh_\nu^\top \mathcal H^{-1}_{[\nu]}\bh_\nu.
$$
This is a heuristic argument; we mark the assumptions used below explicitly. We have
\begin{align*}
    \mathcal{L}_{[\nu]} &= \frac{1}{n}\sum_{\mu \neq \nu}\qty(L(\bgamma^\top\bh_\mu) - y_\mu\bgamma^\top\bh_\mu) + \frac{\lambda d}{2}\|\bgamma\|^2 \\
    \frac{\text{d}\mathcal{L}_{[\nu]}}{\text{d}\bgamma} &= \frac{1}{n}\sum_{\mu \neq \nu}\qty(L'(\bgamma^\top \bh_\mu)-y_\mu)\bh_\mu + {\lambda d}\bgamma \\
    \mathcal H_{[\nu]} &= \frac{1}{n}\sum_{\mu\neq\nu}L''(\fitlogit{\mu,[\nu]})\bh_\mu \bh_\mu^\top + {\lambda d}I_{d(d+1)}
\end{align*}

As in the quadratic expansion, we evaluate $$\mathcal H_{[\nu]} := \frac{\text{d}^2 \mathcal{L}_{[\nu]}}{\text{d}\bgamma^2}\qty(\bgamma = \bgamma^*_{[\nu]}).$$

Eventually we will compare this to the Hessian of the full loss. We have two choices of where to evaluate such an object, at $\bgamma = \bgamma^*$ or at $\bgamma=\bgamma^*_{[\nu]}$. In the latter, they differ by a rank-one term
\begin{align*}
    \mathcal H (\bgamma^*_{[\nu]}) &:=   \frac{\text{d}^2\mathcal{L}}{\text{d}\bgamma^2}\qty(\bgamma = \bgamma^*_{[\nu]}) = \mathcal H_{[\nu]} + \frac{1}{n}L''(\fitlogit{\nu,[\nu]})\bh_\nu\bh_\nu^\top.
\end{align*}
For the former, we must bound $\fitlogit{\mu} - \fitlogit{\mu,[\nu]}$ induced differences, as \begin{align*}
        \mathcal H(\bgamma^*) &:=   \frac{\text{d}^2\mathcal{L}}{\text{d}\bgamma^2}\qty(\bgamma = \bgamma^*) = \mathcal H_{[\nu]} + \frac{1}{n}L''(\fitlogit{\nu})\h_\nu\h_\nu^\top + \frac{1}{n}\sum_{\mu\neq\nu} \qty(L''(\fitlogit{\mu})-L''(\fitlogit{\mu,[\nu]}))\bh_\mu\bh_\mu^\top.
\end{align*}

We assume, without proof, that the matrix remainder
$$
\frac{1}{n}\sum_{\mu\neq\nu}
\qty(L''(\fitlogit{\mu})-L''(\fitlogit{\mu,[\nu]}))
\bh_\mu\bh_\mu^\top
$$
can be neglected in the normalised inverse-Hessian quadratic forms
and traces used below. Expand \begin{align*}
     \mathcal H(\bgamma^*)^{-1} &:= \mathcal H^{-1} \approx\qty(\mathcal H_{[\nu]} + \frac{1}{n}L''(\fitlogit{\nu})\h_\nu\h_\nu^\top)^{-1} \\
     &= \mathcal H_{[\nu]}^{-1} - \frac{1}{n}\frac{L''(\fitlogit{\nu})}{1+ L''(\fitlogit{\nu})s_\nu} \mathcal H_{[\nu]}^{-1}\h_\nu\h_\nu^\top  \mathcal H_{[\nu]}^{-1} \\
     \text{so}\quad\frac{1}{n}\h_{\nu}^\top \mathcal H^{-1} \h_\nu &\approx \frac{s_\nu}{1+L''(\fitlogit{\nu})s_\nu}.
 \end{align*}

We have
\begin{align*}
    I &= \mathcal H^{-1} \mathcal H = \lambda d \mathcal H^{-1} + \frac{1}{n}\sum_{\mu}L''(\fitlogit{\mu})\mathcal H^{-1}\bh_\mu\bh_\mu^\top
\end{align*} and so $$\frac{1}{n}\sum_{\mu=1}^n \frac{1}{1+L''(\fitlogit{\mu})s_\mu} - \frac{\lambda d}{n}\tr(\mathcal H^{-1}) \approx 1-\frac{1}{\tau}.$$

We can simplify the $s_\mu$ quadratic forms as follows. Recall the structure of $\bh = \vecop(\bm{x}_\text{test}\bm{u}^\top)$ and so for any matrix $A \in \mathbb{R}^{d(d+1)\times d(d+1)}$, we have $$\frac{1}{n}\h^\top A\h = \frac{1}{n}x_iu_a A_{ia,jb}x_ju_b.$$ The $\x$s are mean-zero, sub-Gaussian, with independent coordinates, and so if we write matrices $B_{ij} = u_a A_{ia,jb} u_b$ then $$\frac{1}{n}\h^\top A\h \approx \frac{1}{n}\tr(B\E[\x\x^\top]) = \frac{1}{n}\tr(B).$$ For the $\bm{u}$ terms, conditional on $\task$, \begin{align*}
        \E[\bm{u}^\top  \mid \task] &= \begin{bmatrix}
            d\mathcal{C}_\beta\task^\top & 1/2
        \end{bmatrix} \\
       \E\left[\bm{u}\bm{u}^\top \mid \task \right] &=
\begin{bmatrix}
\displaystyle
d^2\mathcal{C}_\beta^2\task\task^\top
+
\frac{d}{\alpha_\mathrm{train}}
\left(
\frac{1}{2}I_d
-
\mathcal{C}_\beta^2\task\task^\top
\right)
&
\displaystyle
\left(
\frac{d\,\mathcal{C}_\beta}{2}
+
\frac{\mathcal{C}_\beta}{2\alpha_\mathrm{train}}
\right)\task
\\[4mm]
\displaystyle
\left(
\frac{d\,\mathcal{C}_\beta}{2}
+
\frac{\mathcal{C}_\beta}{2\alpha_\mathrm{train}}
\right)\task^\top
&
\displaystyle
\frac{1}{4}
+
\frac{1}{4\alpha_\mathrm{train} d}
\end{bmatrix}
\end{align*}
where $$\mathcal{C}_\beta := \E_{g\sim\mathcal{N}(0,1)}[g\sigma(\beta g)].$$
Averaging over the $\|\task\|=1$ spherically uniform task distribution gives \begin{align}\label{eq:leave_one_out_resolvent_hessian}
    \E[\bm{u}^\top] &= \begin{bmatrix}
        \bm{0} & 1/2
    \end{bmatrix} \\
    \E[\bm{u}\bm{u}^\top] &\approx \begin{bmatrix}
        d\qty(\mathcal{C}_\beta^2 + \frac{1}{2\alpha_\mathrm{train}})I_d & \bm{0} \\
        \bm{0}^\top & \frac{1}{4}
    \end{bmatrix} := \Sigma_\beta
\end{align}
For the leave-one-out inverse Hessians considered below, we assume $$\frac{1}{n}\h^\top A \h \approx \frac{1}{n}\tr(A(I_d\otimes \Sigma_\beta)).$$

So we will write $$s_\nu \approx \frac{1}{n}\tr(\mathcal H_{[\nu]}^{-1}(I_d\otimes \Sigma_\beta)).$$

Now we assume that $\mathcal{L}, \mathcal{L}_{[\nu]}$ are sufficiently well behaved so that the relevant normalised traces of the inverse Hessians at their respective minimisers are comparable, i.e.
$$
\frac{1}{n}\tr\qty(\mathcal H_{[\nu]}^{-1}\Sigma_h)
\approx
\frac{1}{n}\tr\qty(\mathcal H^{-1}\Sigma_h).
$$

Thus for $\Sigma_h = I_d \otimes \Sigma_\beta$, we write
$$
s_\nu \approx \frac{1}{n}\tr(\mathcal H_{[\nu]}^{-1}\Sigma_h)
\approx \frac{1}{n}\tr(\mathcal H^{-1}\Sigma_h) := s.
$$
We further assume that this quantity concentrates to a deterministic scalar $s$. We then have $$\frac{1}{n}\sum_{\mu=1}^n \frac{1}{1+L''(\fitlogit{\mu})s} - \frac{\lambda d }{n}\tr(\mathcal H^{-1}) = 1-\frac{1}{\tau}.$$

Estimating $\tr(\mathcal H^{-1})$ will be difficult and we wish to avoid it. For $\lambda>0$, we make the additional approximation, without proof, that the contribution of the final vector block of $\Sigma_\beta$ to the normalised inverse-Hessian trace is asymptotically negligible, so that the main block dominates. Thus,
$$
s \approx \frac{1}{n}\tr(\mathcal H^{-1}(I_d \otimes \Sigma_\beta))
\approx \frac{d\Delta}{n}\tr(\mathcal H^{-1}).
$$
For $\lambda=0$, this approximation is not required, since the term involving $\tr(\mathcal H^{-1})$ in the preceding equation vanishes.

To summarise, we have that $$s_\nu = \frac{1}{n}\h_\nu^\top \mathcal H^{-1}_{[\nu]}\h_\nu \approx s$$ for deterministic scalar $s$ defined self-consistently by \begin{align*}
    1 - \frac{1}{\tau} &= \E\qty[\frac{1}{1+L''(R_s)s}] - \frac{\lambda}{\Delta}s.
\end{align*}
In the limit as $\lambda \to 0$, we have the nicer ridgeless characterisation
$$1 - \frac{1}{\tau} = \E\qty[\frac{1}{1+L''(R_s)s}].$$

\paragraph{Step 3. Gaussian equivalent} We will construct a logit that is statistically equivalent to the left-out logits $R_\mathrm{loo}$. To do this, we make a standard assumption commonly used in cavity arguments, that asymptotically there is no difference between $\h_\mu^\top\bgamma_{[\mu]}^*$ and $\h_\text{new}^\top\bgamma^*$. Moreover, $\bgamma_{[\mu]}^*$ is independent of $\h_\mu$, since prompt $\mu$ is excluded from its training set. Here $\h_\text{new}$ is drawn from the same distribution as the training distribution, namely context length is still $\ell$ and is not $\ell_\text{test}$. This distinction is important, as $\h_\text{new}^\top\bgamma^*$ here is not the test logit, which would describe sequences of length $\ell_\text{test}$.

Thus we have that
\[
    R_\mathrm{loo} \overset{d}{=}
    \x_q^\top
    \left(
        \vA^* \bm{v} + c \bm{a}^*
    \right),
    \qquad
    \bm{v}
    :=
    \frac{d}{\ell}\sum_{i=1}^{\ell} y_i\x_i,
    \qquad
    c
    :=
    \frac{1}{\ell}\sum_{i=1}^{\ell} y_i.
\]
Write
\[
    \vA^*
    =
    \frac{\hataone}{d}I_d+\bar{\vA}^*,
    \qquad
    \hataone:=\tr(\vA^*),
    \qquad
    \tr(\bar{\vA}^*)=0.
\]
Then
\[
    R_{\mathrm{loo}}
    =
    \frac{\hataone}{d}\x_q^\top\bm{v}
    +
    \x_q^\top\bar{\vA}^*\bm{v}
    +
    c\,\x_q^\top\bm{a}^*.
\]
We analyse these three terms separately.

We make a standard cavity self-averaging assumption: conditional on the leave-one-out estimator, the relevant overlaps and quadratic forms of a fresh prompt concentrate around some deterministic limits. Since $\x_q$ is Gaussian and independent of context variables and the leave-one-out estimator, the terms below are conditionally jointly Gaussian. As their conditional covariances concentrate, their limiting joint law is derived by the corresponding covariance calculations below.

First, write
\[
    \x_q=G_q\task+\x_\text{perp},
    \qquad
    G_q:=\x_q^\top\task\sim\cN(0,1),
    \qquad
    \x_\text{perp}\sim\cN(0,I_d-\task\task^\top).
\]
Conditional on \(\task\), we have
\[
    \E[\bm{v}\mid\task]=d\mathcal{C}_\beta\task
\]
and
\[
    \text{Cov}(\bm{v}\mid\task)
    =
    \frac{d}{2\alpha_\mathrm{train}}I_d
    -
    \frac{d\mathcal{C}_\beta^2}{\alpha_\mathrm{train}}\task\task^\top.
\]
Consequently,
\[
    \frac{1}{d}\x_q^\top\bm{v}
    =
    \mathcal{C}_\beta G_q
    +
    \frac{1}{d}\x_q^\top
    \left(
        \bm{v}-d\mathcal{C}_\beta\task
    \right).
\]
Using concentration of quadratic forms,
\[
    \text{Var}\left(
        \frac{1}{d}\x_q^\top
        \left(
            \bm{v}-d\mathcal{C}_\beta\task
        \right)
    \right)
    \approx
    \frac{1}{2\alpha_\mathrm{train}}.
\]

Moreover, its covariance with \(G_q\) is asymptotically zero. Under the self-averaging assumption above, it results in
\[
    \frac{\hataone}{d}\x_q^\top\bm{v}
    \ \overset{d}{\approx}\
    \hataone\mathcal{C}_\beta G_q
    +
    \frac{\hataone}{\sqrt{2\alpha_\mathrm{train}}}G_{\tr},
\]
where \(G_{\tr}\sim\cN(0,1)\) is independent of \(G_q\).

We next consider the traceless contribution. Since
\[
    \tr(\bar{\vA}^*)=0,
\]
we have
\[
    \E\left[
        \x_q^\top\bar{\vA}^*\bm{v}
    \right]
    \approx
    d\mathcal{C}_\beta
    \E\left[
        \x_q^\top\bar{\vA}^*\task
    \right]
    =0.
\]
We have
\begin{align*}
    \E\left[
        \left(
            \x_q^\top\bar{\vA}^*\bm{v}
        \right)^2
    \right]
    &\approx
    d^2\mathcal{C}_\beta^2
    \E\left[
        \left(
            \x_q^\top\bar{\vA}^*\task
        \right)^2
    \right] +
    \frac{d}{2\alpha_\mathrm{train}}
    \E\left[
        \x_q^\top
        \bar{\vA}^*
        \bar{\vA}^{*\top}
        \x_q
    \right].
\end{align*}
Averaging over \(\x_q\) and $\|\task\|^2 = 1$ gives
\[
    \E\left[
        \left(
            \x_q^\top\bar{\vA}^*\task
        \right)^2
    \right]
    =
    \frac{1}{d}
    \|\bar{\vA}^*\|_F^2
\]
and
\[
    \E\left[
        \x_q^\top
        \bar{\vA}^*
        \bar{\vA}^{*\top}
        \x_q
    \right]
    =
    \|\bar{\vA}^*\|_F^2.
\]
Therefore
\[
    \E\left[
        \left(
            \x_q^\top\bar{\vA}^*\bm{v}
        \right)^2
    \right]
    \approx
    d\|\bar{\vA}^*\|_F^2
    \left(
        \mathcal{C}_\beta^2+\frac{1}{2\alpha_\mathrm{train}}
    \right).
\]
Note that
\begin{align*}
    d\|\bar{\vA}^*\|_F^2 =
    \hatatwo-\hataone^2
\end{align*}
and so
\[
    \x_q^\top\bar{\vA}^*\bm{v}
    \ \overset{d}{\approx}\
    G_{\mathrm{traceless}}
    \sqrt{
        (\hatatwo-\hataone^2)
        \left(
            \mathcal{C}_\beta^2+\frac{1}{2\alpha_\mathrm{train}}
        \right)
    }.
\]

Finally, since \(c\approx 1/2\), we have
\[
    c\,\x_q^\top\bm{a}^*
    \ \overset{d}{\approx}\
    G_{\mathrm{vec}}\sqrt{\frac{\hatazero}{4}},
    \qquad
    \hatazero:=\|\bm{a}^*\|^2.
\]
Combining everything gives a joint statistically-equivalent characterisation \begin{align*}
    R_\mathrm{loo} \overset{d}{=} \hataone\mathcal{C}_\beta G_\text{q} + \hataone \frac1{\sqrt{2\alpha_\mathrm{train}}}G_\text{tr}
+
G_\text{traceless}\sqrt{(\hatatwo-\hataone^2)
\qty(
\mathcal{C}_\beta^2+\frac1{2\alpha_\mathrm{train}}
)
}
+
G_\text{vec} \sqrt{\frac{\hatazero}{4}},
\end{align*} where $$G_\text{q}, G_\text{tr}, G_\text{traceless}, G_\text{vec} \sim_\text{i.i.d.} \mathcal{N}(0,1).$$

\paragraph{Step 4. Closing the proximal equation} We now assume that the empirical order parameters $(\hataone,\hatatwo,\hatazero)$ concentrate around the deterministic limits $(\aone,\atwo,\azero)$. We also assume that $s_\mu\to s$. Exchangeability then identifies the limiting joint law of $(\fitlogit{\mu,[\mu]},\fitlogit{\mu},y_\mu)$ with $(R_{\mathrm{loo}},R_s,Y)$, where $R_{\mathrm{loo}}=R_s+s(\sigma(R_s)-Y)$. From this point onward we use the scalar variables $(R_{\mathrm{loo}},R_s,Y)$ after converting empirical averages to expectations. The proof here will proceed in a heuristic fashion. For $\lambda \to 0$, we assume without proof that the limiting order parameters are $O(1)$ with respect to $d$; we have numerically validated this. For $\lambda \neq 0$, additional control of the scaling of the order parameters is required, so we retain the corresponding terms in the final expression.

To proceed, we can differentiate $\mathcal{L}$ with respect to our desired parameters to get \begin{align*}
    \frac{\partial \mathcal{L}}{\partial \hataone}&=\frac{1}{n}\sum_\mu(\sigma(\fitlogit{\mu})-y_\mu)\frac{\x_{\text{q},\mu}^\top\bm{v}_\mu}{d} + \lambda \hataone = 0 \\
    \frac{\partial \mathcal{L}}{\partial \bm{a}}&=\frac{1}{n}\sum_\mu(\sigma(\fitlogit{\mu})-y_\mu)c_\mu \x_{\text{q},\mu}+\lambda d \bm{a} = \bm{0}.
\end{align*}
We focus first on the $\hataone$ equation. For $\lambda=0$ this expression simplifies; for general $\lambda$, we retain the $\lambda\hataone$ term. Using the proximal equation, concentration $\hataone\to\aone$, and approximating $1/n\sum_\mu$ by an
expectation, we have
\begin{equation*}
    \E\qty[(R_{\mathrm{loo}}-R_s)\frac{1}{d}\x_\text{q}^\top\bm{v}]
    +s{\lambda}\aone=0.
\end{equation*}
We can expand \begin{align*}
    \E[R_{\mathrm{loo}}\frac{1}{d}\x_\text{q}^\top\bm{v}] &= \frac{1}{d}\E\qty[\tr\qty((A\bm{v}\bm{v}^\top + \bm{a}(c\bm{v})^\top)\x_\text{q}\x_\text{q}^\top)] \\
    &= \frac{1}{d}\E\qty[\tr\qty(A\bm{v}\bm{v}^\top + \bm{a}(c\bm{v})^\top)] \\
    &= \aone\qty(\mathcal{C}_\beta^2 + \frac{1}{2\alpha_\mathrm{train}}).
\end{align*}  Furthermore, using the same tricks as previously, we can write the marginal for $\x_\text{q}^\top\bm{v}$ as $$\frac{1}{d}\x_\text{q}^\top\bm{v} \overset{d}{=} \mathcal{C}_\beta G_\text{q} + \frac{1}{\sqrt{2\alpha_\mathrm{train}}}G_\text{tr}.$$ Thus
\[
 \aone = \frac{\E\qty[R_s\qty(\mathcal{C}_\beta G_\text{q}+ \frac{1}{\sqrt{2\alpha_\mathrm{train}}}G_\text{tr})]}{\Delta + s \lambda }.
\]

Now for $\bm{a}$, set
\[
    W:=c\,\bm{a}^{*\top}\x_\mathrm{q}.
\]

The vector first-order condition and the proximal equation give
\begin{align*}
    0
    &=\E\big[(\sigma(R_s)-Y)W\big]+\lambda d \hatazero \\
    &=\frac1s\E\big[(R_{\mathrm{loo}}-R_s)W\big]+\lambda d \hatazero.
\end{align*}
Thus
\[
    \E[R_{\mathrm{loo}}W]-\E[R_s W]+s\lambda d \hatazero=0.
\]
Under the Gaussian equivalent above, $W\overset{d}{\approx}\sqrt{\hatazero/4}\,G_\mathrm{vec}$ and
$G_\mathrm{vec}$ is independent of the other cavity variables. Hence Stein's lemma gives
\[
    \E[R_{\mathrm{loo}}W]\approx\frac{\hatazero}{4},
    \qquad
    \E[R_s W]\approx\frac{\hatazero}{4}\chi(s),
\]
where $\chi(s)=\E[(1+s\sigma'(R_s))^{-1}]$. Therefore, asymptotically,
\[
    \hatazero\left(\frac{1-\chi(s)}{4}+s\lambda d \right)\longrightarrow 0.
\]
This forces $\azero=0$ and, moreover, $s\lambda d\hatazero\to0$.

From the full derivative equation for $\bgamma$ we have
\[
    \E[(\sigma(R_s)-Y)R_s]
    +\lambda d \qty(\frac{\hatatwo}{d}+\hatazero)=0.
\]
Using the proximal equation, this becomes
\[
    \E[R_{\mathrm{loo}}R_s]-\E[R_s^2]
    +s\lambda d \qty(\frac{\hatatwo}{d}+\hatazero)=0.
\]
Taking the limit and using $s\lambda d\hatazero\to0$ gives
\[
    \E[R_{\mathrm{loo}}R_s]-\E[R_s^2]+s\lambda\atwo=0.
\]

In the ridgeless case, the final term vanishes. For finite ridge, we have the following self-consistency equation after using $\azero=0$:
\[
 \atwo=\frac{\E[R_s^2]-\Delta(1-\chi)\aone^2 - s\lambda \aone^2}
 {\Delta\chi+s\lambda}.
\]

In the explicitly ridgeless case, we may expand $\E[R_{\mathrm{loo}}R_s]$ at leading order, giving
\begin{align*}
    \E[R_{\mathrm{loo}}R_s] &= \aone\E\qty[R_s\qty(\mathcal{C}_\beta G_\text{q} + \frac1{\sqrt{2\alpha_\mathrm{train}}}G_\text{tr})]
+
\E\qty[R_s G_\text{traceless}\sqrt{(\atwo-\aone^2)\qty(\mathcal{C}_\beta^2+\frac1{2\alpha_\mathrm{train}})}] \\
&= \aone^2\qty(\mathcal{C}_\beta^2 + \frac{1}{2\alpha_\mathrm{train}}) + (\atwo-\aone^2)\qty(\mathcal{C}_\beta^2+\frac1{2\alpha_\mathrm{train}})\E\qty[\frac{\partial R_s}{\partial R_{\mathrm{loo}}}] \quad \text{by Stein's lemma}.
\end{align*} Now $$\frac{\partial R_s}{\partial R_{\mathrm{loo}}} = \frac{1}{1+ s\sigma'(R_s)}$$ from differentiating the proximal equation, and so $$\atwo = \qty(1-1/\E\qty[\frac{1}{1+ s\sigma'(R_s)}])\aone^2 + \frac{\E[R_s^2]}{\qty(\mathcal{C}_\beta^2+\frac1{2\alpha_\mathrm{train}})\E\qty[\frac{1}{1+ s\sigma'(R_s)}]}.$$

\paragraph{Step 5. Characterisation of the test logit} Here we apply a similar argument to one used in Step 3. We wish to characterise the test logit $\h_\mathrm{test}^\top\bgamma^*$. Thus,
\[
    R_\mathrm{test} =
    \x_q^\top
    \left(
        \vA^* \bm{v} + c \bm{a}^*
    \right),
    \qquad
    \bm{v}
    :=
    \frac{d}{\ell_\text{test}}\sum_{i=1}^{\ell_\text{test}} y_i\x_i,
    \qquad
    c
    :=
    \frac{1}{\ell_\text{test}}\sum_{i=1}^{\ell_\text{test}} y_i.
\]
The argument for a statistically equivalent characterisation proceeds exactly the same as in Step 3, except the moments involve $\alpha_\mathrm{test}$ now instead of $\alpha_\text{train}$. This gives the desired marginal distribution
 \begin{align*}
    y_\text{q} &\sim \text{Bernoulli}(\sigma(\beta G_\text{q})) \\
    R_\mathrm{test} &= \aone \mathcal{C}_\beta G_\text{q} + \aone \frac1{\sqrt{2\alpha_\mathrm{test}}}G_\text{tr}
+
G_\text{traceless}\sqrt{(\atwo-\aone^2)\Delta_\mathrm{test}}, \quad \Delta_\mathrm{test}
:=\mathcal C_\beta^2+\frac{1}{2\alpha_\mathrm{test}}.
\end{align*} This concludes the proof of Result \ref{lemma:a1a2eqs} and Result \ref{prop:main_result_ridgeless}.

\end{proof}

\section{Schematic derivation of solution boundary Result \ref{lemma:critical_tau}}\label{sec:informal_solution_boundary_proof}
This claim follows by applying the result of \cite{koriyama2025phasetransitionsexistenceunregularized} to our problem setting, which requires an additional heuristic transfer argument. The setting considered by \cite{koriyama2025phasetransitionsexistenceunregularized} is the classic teacher-student setting where the model being trained has Gaussian covariates $\x$. This assumption clearly does not hold in our setting, where not only do the model's covariates $\h$ not match the teacher covariates $\x$, but the $\h$s are not Gaussian. That said, we present the following schematic argument for the applicability of their result to our setting: each $\h$ takes the form $$\h=\vecop(H), \quad H = \x_\mathrm{q}\bm{u}^\top, \quad \bm{u}^\top = \qty[\frac{d}{\ell}\sum_{i=1}^\ell y_i\x_i^\top\;\; \frac{1}{\ell}\sum_{i=1}^{\ell}y_i^2].$$ We see that $\x_\mathrm{q}$ is Gaussian, and $\bm{u}$ is nearly Gaussian (conditional on $\task$), allowing us to exploit the tensorised form of $\h$ to transfer the result of \cite{koriyama2025phasetransitionsexistenceunregularized} (similar to how we exploited this form in arguing the concentration of quadratic forms $\h^\top R \h$). We stress that this is a heuristic argument.

The second caveat is that the statement in Result \ref{lemma:critical_tau} is quite informal. The correct statement is $$\mathbb{P}(\text{finite solution to ridgeless problem exists}) \to \begin{cases}
    0 & \tau < \tau_c \\
    1 & \tau > \tau_c
\end{cases}.$$

Given the above, all that remains is to directly apply the result. We take $G\sim \mathcal{N}(0,1)$ independently, and define $(U,Y)$ by the limiting joint distribution of
$(\x_\mathrm{q}^\top\bm{v}/(d\sqrt{\Delta}),y_\mathrm{q})$. We know from the previous marginal characterisation that $$U \propto \mathcal{C}_\beta G_\mathrm{q} + \frac{1}{\sqrt{2\alpha_\mathrm{train}}}G_\mathrm{tr}$$ where $G_\mathrm{q},G_\mathrm{tr}\sim\mathcal{N}(0,1)$ with $Y \sim \text{Bernoulli}(\sigma(\beta G_\mathrm{q}))$. Normalising the variance of $U$ gives the required result.

\section{Properties of loss functions with context length}\label{sec:test_time_monotonic}
Here we justify our claim that increasing test-time context length strictly decreases both losses. Intuitively, this follows because $\alpha_\mathrm{test}$ only appears in the variance term $V$ that contributes randomness that is independent of the actual signal $G_\mathrm{q}$.

To be concrete, for the cross-entropy loss, we can write \begin{align*}
    \mathcal{E}_\mathrm{CE} &= \E[L(R)] - \E[\sigma(\beta G_\mathrm{q})a_1\mathcal{C}_\beta G_\mathrm{q}] \\
    &= \E_{G\sim\mathcal{N}(0,1)}\qty[\log(1 + \exp(cG))] - a_1 \mathcal{C}_\beta^2
\end{align*} where $$c = \sqrt{a_2\qty(\frac{1}{2\alpha_\mathrm{test}}+\mathcal{C}_\beta^2)}$$ does not depend on $G$. Thus $$\frac{\partial\mathcal{E}_\mathrm{CE}}{\partial c} = \E\qty[\frac{G\exp(cG)}{1+\exp(cG)}] = \E\qty[\frac{c\exp(cG)}{(1+\exp(cG))^2}] \geq 0$$ and the result follows as $\partial c /\partial \alpha_\mathrm{test} < 0$.

For the classification loss, write $$\mathcal{E}_\mathrm{class} = \E[\sigma(\beta G_\mathrm{q})] + \E[(1-2\sigma(\beta G_\mathrm{q}))\phi]$$
where again $$\phi = \Phi\qty(sG_\mathrm{q}),\qquad s = \frac{a_1\mathcal{C}_\beta}{\sqrt{\frac{a_2}{2\alpha_\text{test}} + (a_2 -a_1^2)\mathcal C_\beta^2}}.$$ Thus $\phi$ is a decreasing function in $s$ for $G < 0$ and increasing in $s$ for $G > 0$. We see that $s$ is an increasing function of $\alpha_\text{test}$, and so $\phi$ is decreasing in $\alpha_\text{test}$ for $G < 0$ and increasing in $\alpha_\text{test}$ for $G > 0$.   These cases exactly correspond to $1-2\sigma(\beta G_\mathrm{q}) > 0$ and $1-2\sigma(\beta G_\mathrm{q}) < 0$, and so $\mathcal{E}_\mathrm{class}$ is always a decreasing function of $\alpha_\mathrm{test}.$

\end{document}